\documentclass[10pt,twocolumn,letterpaper]{article}

\usepackage{algorithm}
\usepackage{algpseudocode}
\usepackage{amsmath}
\usepackage{amssymb}
\usepackage{array}
\usepackage{booktabs}
\usepackage{caption}
\usepackage{cite}
\usepackage{color}
\usepackage{cvpr}
\usepackage{dirtytalk}
\usepackage{epsfig}
\usepackage{graphicx}
\usepackage{gensymb}
\usepackage{lipsum}
\usepackage{makecell}
\usepackage{mathtools}
\usepackage{multirow}
\usepackage{stmaryrd}
\usepackage{times}
\usepackage{xcolor}
\usepackage{xcolor}

\usepackage[position=bottom]{subfig}
\usepackage[font=small,labelfont=bf,tableposition=top]{caption}

\algnewcommand\algorithmicforeach{\textbf{for each}}
\algdef{S}[FOR]{ForEach}[1]{\algorithmicforeach\ #1\ \algorithmicdo}

\usepackage[breaklinks=true,bookmarks=false,urlcolor=blue,colorlinks]{hyperref}

\usepackage[utf8]{inputenc}
\DeclareUnicodeCharacter{2212}{-}

\DeclareMathOperator*{\argmin}{arg\,min}

\newcommand{\ra}[1]{\renewcommand{\arraystretch}{#1}}
\def\m#1{\ensuremath{\mathtt{#1}}}

\def\v#1{\ensuremath{\mathbf{#1}}}

\def\norm#1{\left\lVert#1\right\rVert}
\def\parr#1{\left(#1\right)}
\def\curl#1{\left\{#1\right\}}
\def\croch#1{\left[#1\right]}
\def\iverson#1{\left\llbracket#1\right\rrbracket}

\def\I{\m I}
\def\Q{\m Q}

\def\H{\m H}
\def\h{\v h}
\def\R{\m R}
\def\G{\m G}
\def\t{\v t}

\def\S{\m S}
\def\C{\m C}
\def\Ctilde{\tilde{\m C}}
\def\L{\m L}

\def\K{\m K}
\def\p{\v p}
\def\q{\v q}

\def\r{\v r}

\def\u{\v u}

\def\T{\m T}

\def\H{\m H}

\def\out{\v{out}}

\newcommand{\defeq}{\vcentcolon=}

\definecolor{orange}{rgb}{1.0,0.5,0}

\definecolor{DarkGreen}{rgb}{0,0.5,0}

\newcommand{\gbrmk}[1]{}
\newcommand{\gbrm}[1]{}

\newcommand{\vincentrmk}[1]{}

\newcommand{\hugormk}[1]{}

\def\HideSuppMat{0}

\cvprfinalcopy 
\definecolor{pink}{RGB}{255,192,203}
\definecolor{paleturquoise}{RGB}{175,238,238}
\definecolor{lightsteelblue}{RGB}{176,196,222}
\definecolor{violet}{RGB}{238,130,238}
\definecolor{orange}{RGB}{255,120,0}
\definecolor{mediumpurple}{RGB}{147,112,219}
\definecolor{lightgreen}{RGB}{0,128,0}
\definecolor{purple}{RGB}{128,0,128}

\def\cvprPaperID{2202} 

\begin{document}

\title{Neural Reprojection Error: Merging Feature Learning and \\Camera Pose Estimation} 

\author{
  Hugo Germain$^1$ \qquad Vincent Lepetit$^1$ \qquad Guillaume Bourmaud$^2$  \\
  $^1$LIGM, \'Ecole des Ponts, Univ Gustave Eiffel, CNRS, Marne-la-vall\'ee, France\\
  $^2$IMS, University of Bordeaux, Bordeaux INP, CNRS, France\\
  {\tt\small \{hugo.germain, vincent.lepetit\}@enpc.fr \qquad guillaume.bourmaud@u-bordeaux.fr}
}


\maketitle

\begin{abstract}
Absolute camera pose estimation is usually addressed by sequentially solving two distinct subproblems: First a feature matching problem that seeks to establish putative 2D-3D correspondences, and then a Perspective-n-Point problem that minimizes, \wrt the camera pose, the sum of so-called Reprojection Errors (RE). We argue that generating putative 2D-3D correspondences 1) leads to an important loss of information that needs to be compensated as far as possible, within RE, through the choice of a robust loss and the tuning of its hyperparameters and 2) may lead to an RE that conveys erroneous data to the pose estimator. In this paper, we introduce the Neural Reprojection Error (NRE) as a substitute for RE. NRE allows to rethink the camera pose estimation problem by merging it with the feature learning problem, 
hence leveraging richer information than 2D-3D correspondences and eliminating the need for choosing a robust loss and its hyperparameters.
Thus NRE can be used as training loss to learn image descriptors tailored for pose estimation. We also propose a coarse-to-fine optimization method able to very efficiently minimize a sum of NRE terms \wrt the camera pose.  We experimentally demonstrate that NRE is a good substitute for RE  as it significantly improves both the robustness and the accuracy of the camera pose estimate while being computationally and memory highly efficient.
From a broader point of view, we believe this new way of merging deep learning
and 3D geometry may be useful in other computer vision applications. 
Source code and model weights will be made available at \href{https://www.hugogermain.com/nre}{hugogermain.com/nre}.

\end{abstract}


\begin{figure}
  \begin{center}
    \centering
    \includegraphics[width=\columnwidth]{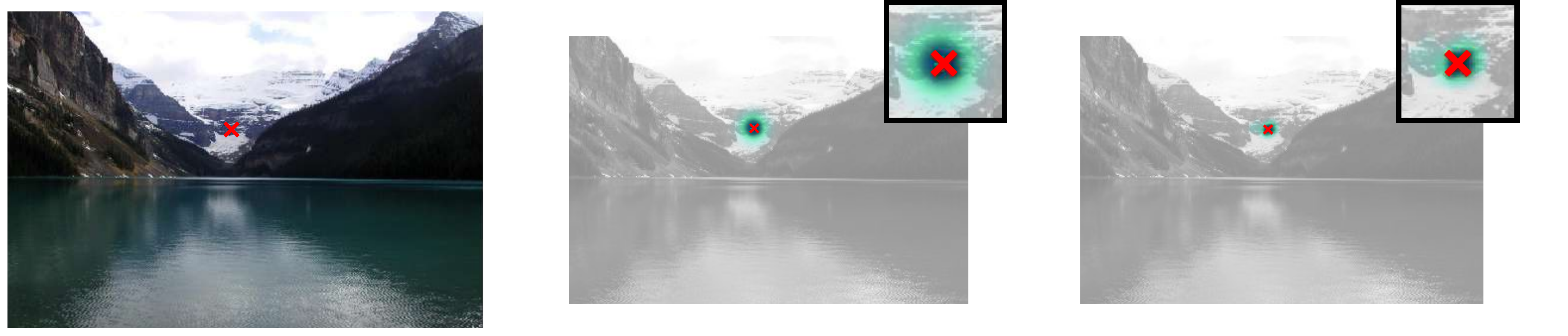}
    \includegraphics[width=\columnwidth]{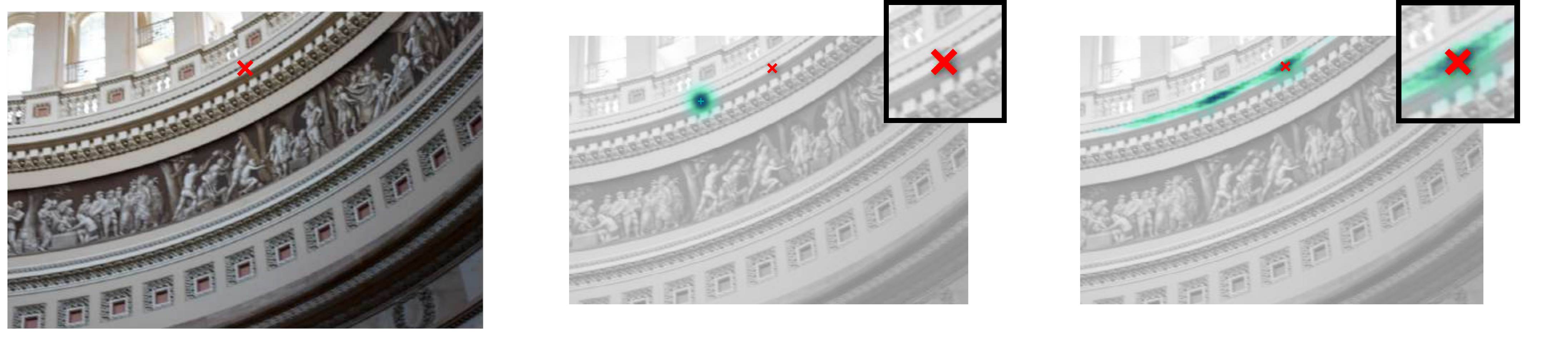}
    \includegraphics[width=\columnwidth]{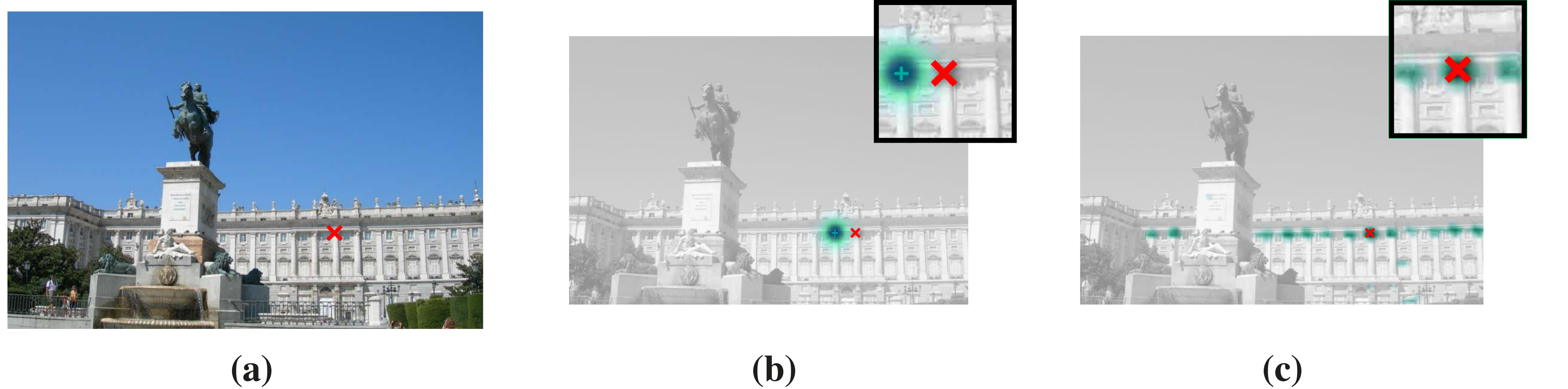}
    \caption{
      \textbf{Neural Reprojection Error (NRE) as a substitute for Reprojection Error (RE):} \textbf{(a)} Given a 3D point $\u$, a query image $\I$ and its ground truth camera pose, $\u$ 
can be reprojected into the image plane of $\I$ to obtain a 2D point $\color{red}\boldsymbol{\times}$. \textbf{(b)} RE takes as input a camera pose and a putative 2D-3D correspondence between $\u$ and a 2D location $\color{DarkGreen}\boldsymbol{+}$ in $\I$, reprojects $\u$ to obtain a 2D point $\q$, computes the euclidean distance between $\color{DarkGreen}\boldsymbol{+}$ and $\q$ and finally applies a robust loss function (shown in turquoise as a function of $\q$). 
In ambiguous (middle) or multimodal (bottom) cases, generating a 2D-3D correspondence may lead to a loss function that conveys erroneous data to the pose estimator. \textbf{(c)} NRE does not rely on 2D-3D correspondences, thus $\color{DarkGreen}\boldsymbol{+}$ does not exist anymore. Instead, NRE employs a dense loss map (shown in turquoise as a function of $\q$) that contains much more information than RE, especially in ambiguous and multimodal cases. As a result, a pose estimator is significantly more accurate and robust using NRE than RE.
    }
    \label{fig:teaser}
    \vspace{-0.6cm}
  \end{center}
\end{figure}

%

\section{Introduction}



\begin{figure*}
  \begin{center}
    \centering
    \includegraphics[width=\textwidth]{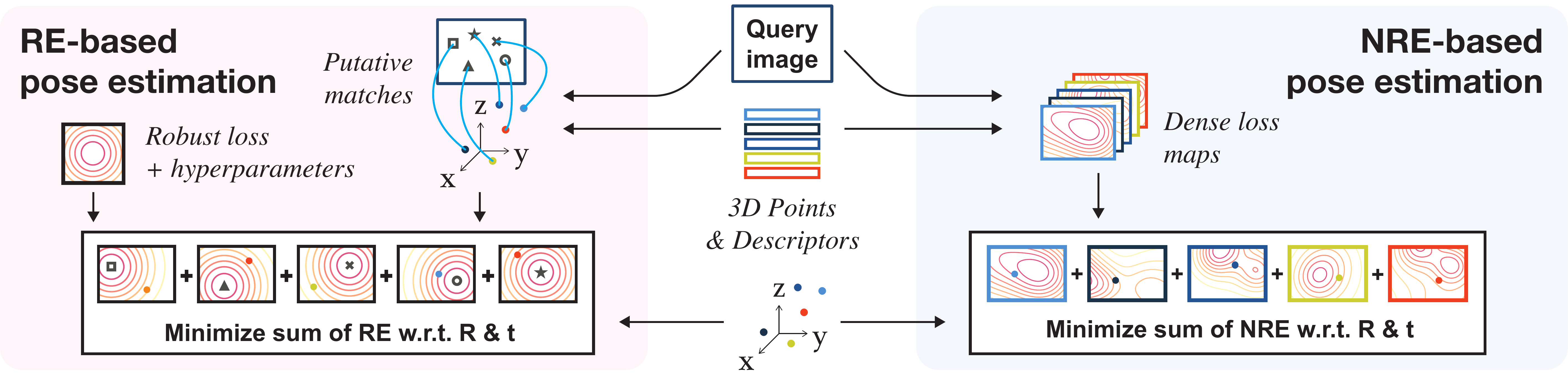}
    \caption{\label{fig:overview} NRE-based pose estimator  \vs RE-based pose estimator.   \textbf{Left:}  In an RE-based pose estimator, putative  2D-3D  correspondences are  initially established.  Then, a sum of RE terms is minimized, \wrt the camera pose, using these correspondences, a robust loss and its hyperparameters as inputs. \textbf{Right:} In an NRE-based pose estimator, dense loss maps are computed instead of 2D-3D correspondences. Then the query pose is estimated by
      minimizing a sum of NRE terms, effectively leveraging
      richer information than simple 2D-3D correspondences and alleviating the need for choosing a robust loss and its hyperparameters.}
      \vspace{-0.3cm} 
  \end{center}
\end{figure*}

Absolute camera pose estimation is a fundamental step to many computer vision applications, such as Structure-from-Motion (SfM)~\cite{Heinly2015, Schoenberger2016sfm, Schoenberger2016mvs, Sweeney2016} and visual localization~\cite{CSL, Svrm2014AccurateLA, Sattler:hal-01513083}.  Given a pre-acquired 3D model of the world, we aim at estimating the most accurate camera pose of an unseen query image.  In practice, as illustrated on the left hand-side of Figure~\ref{fig:overview}, this problem is often addressed by sequentially solving two distinct subproblems: First, a feature matching problem that seeks to establish putative 2D-3D correspondences between the 3D point cloud and the image to be localized, and then a Perspective-n-Point (P\textit{n}P) problem that uses these correspondences as inputs to minimize a sum of so-called reprojection errors \wrt the camera pose. 
The Reprojection Error (RE) is a function of a 2D-3D correspondence and the camera pose.  It consists in reprojecting the 3D point, using the camera pose, into the query image plane, computing the euclidean distance between this reprojection and its putative 2D correspondent, and applying a robust loss function, such as Geman-McClure or Tukey’s biweight~\cite{zach2017iterated,barron2019general}. The robust loss allows to reduce the influence of outlier 2D-3D correspondences.


We argue that this strong decoupling of the matching stage from the P\textit{n}P stage limits both the accuracy and the robustness of the camera pose estimate. Generating putative 2D-3D correspondences leads to an important loss of information since the 3D model and the query image are summarized into a set of 2D-3D coordinates. This loss of information needs to be compensated as far as possible within RE through the choice of a robust loss and the tuning of its hyperparameters, which usually depend on both the visual content and the amount of outliers generated by the matching stage. Moreover, outlier correspondences convey erroneous data to the pose estimator (see fig.~\ref{fig:teaser}).

\noindent\textit{Contributions:}\\
\noindent(i) 
We propose the \emph{Neural Reprojection Error} (NRE) as a substitute for RE. NRE does not require a 2D-3D correspondence as input but relies on a \emph{dense loss map}.  A \emph{dense loss map} contains much more information than a simple 2D-3D correspondence and conveys data of higher quality to the pose estimator. As a result,  the need for choosing a robust loss and its hyperparameters is also eliminated. Computing a \emph{dense loss map} essentially involves cross-correlations between descriptors that are extracted using a neural network, hence the name \emph{Neural Reprojection Error}.

\noindent(ii) Our derivation of NRE makes it differentiable not only \wrt to the camera pose but also \wrt the descriptors. 
Thus, providing ground-truth camera poses and minimizing NRE \wrt the descriptors yields a well-posed feature learning problem tailored for pose estimation. NRE merges the feature learning problem and the camera pose estimation problem in a new way and allows to rethink the recent end-to-end direct feature metric pose refinement methods that need to consider two different losses.

\noindent(iii) To estimate the camera pose efficiently, we propose to minimize a sum of NRE terms in a coarse-to-fine manner. As a result, we never compute or store any high-resolution dense loss map. We also describe how to perform the optimization using an M-estimator sample consensus approach followed by a graduated non-convexity procedure. We experimentally demonstrate that our novel NRE-based pose estimator is a good substitute for RE-based pose estimators as it significantly improves both the robustness and the accuracy of the camera pose estimate while being computationally and memory highly efficient.

In the remainder of the paper, after discussing the related work, we introduce some notations and describe our method. We provide a detailed discussion to highlight the differences between NRE and existing approaches. We finally present our evaluation results.

\section{Related work}
NRE has connections with several research areas, namely, feature learning, learning to match features, end-to-end camera pose estimation and robust optimization. A detailed literature review on these topics seems out of the scope of this paper. Instead, for each topic, we will explain how NRE is related to it and refer the reader to recent papers containing a detailed literature review on it.

\noindent\textbf{Feature learning methods}~\cite{LIFT, R2D2, SuperPoint, D2Net,
  SOSNet, HardNet, Snavely,
  S2DNet,rocco2020efficient,tyszkiewicz2020disk,shen2019rf,benbihi2019elf,bhowmik2020reinforced,luo2020aslfeat}
  learn to transform an image into robust dense descriptors.  Minimizing NRE
  \wrt the descriptors allows to learn features tailored for pose estimation.
  Our training loss is similar to the one proposed in S2DNet~\cite{S2DNet}. Thus
  S2DNet features are in theory well suited to be used as inputs of our novel
  NRE-based pose estimator. However, as we show in our experiments, S2DNet
  computes by nature high-resolution dense correspondence maps which is both
  computationally and memory highly inefficient, hence making our NRE-based pose
  estimator impractical. By merging feature learning and pose estimation, our
  loss intrinsically integrates a bilinear interpolation operator. It allows us
  to learn \emph{coarse} robust features and \emph{fine} discriminative features
  which we combine in a coarse-to-fine strategy. As a result, using our \say{NRE
  features} as input, our NRE-based pose estimator is both fast and memory highly efficient but also significantly more robust and accurate compare to the case where we use S2DNet features as input.

\noindent\textbf{Learning-based matching methods}~\cite{OANet, NGRANSAC,SuperGlue,moo2018learning,sun2020acne,choy2020high} take descriptors and/or putative correspondences as input and output probabilities of correspondences. NRE takes as input dense loss maps which are essentially the negative logarithm of probabilities of correspondences. In our current formulation, we do not employ any sophisticated learning-based matching method to produce these inputs, but a single dot product between descriptors followed by a softmax. Using a state-of-the-art matching architecture would likely improve the results of NRE but we left this as future work.

\noindent\textbf{End-to-end camera pose estimation methods}~\cite{Brachmann2017DSACD,Brachmann2017LearningLI,kendall2015posenet, kendall2017geometric, bui20206d, lv2019taking, tang2018ba, von2020gn} learn jointly all the parameters of the camera pose estimator by backpropagating through it. Different architectures of camera pose estimators have been proposed in the literature.
%
Among these architectures, end-to-end feature metric pose refinement methods \cite{ lv2019taking, tang2018ba, von2020gn}, are the ones that are the closest to NRE as their architectures explicitly minimize a sum of reprojection errors by leveraging richer information than simple 2D-3D correspondences. In Sec.~\ref{sec:discussion-direct-RE} we provide a detailed discussion to explain the fundamental differences between these methods and NRE.

\noindent\textbf{Robust optimization methods}~\cite{barron2019general,DEGENSAC,
MAGSAC, MAGSACpp,barath2018graph,zach2018descending} are tailored to minimize a
sum of non-convex terms. This is essentially what the P\textit{n}P stage seeks
to achieve as it consists in minimizing a sum of RE terms.  In
Sec.~\ref{sec:discussion-RE} we provide a detailed discussion to highlight the
fact that RE is a special case of NRE which allows to relate NRE to standard
robust optimization problems. From another point of view, recent methods, such
as \cite{MAGSAC, MAGSACpp}, allow to eliminate the need for setting the
hyperparameter of the robust loss by marginalizing it. NRE is also able to
eliminate this need  but in a very different manner. Consequently, in the
experiments we will compare the performances of these RE-based estimators
against our novel NRE-based estimator.

\section{Background and notations}\label{sec:background}
In this paper, we assume a sparse 3D point cloud $\left\{ \u_{n}^{\G}\right\} _{n=1...N}$, whose coordinates are expressed in a global coordinate system $\G$, as well as a database $\mathcal{D}$ of geo-localized (\wrt $\G$) reference images are given, and we seek to estimate the pose (\ie the rotation matrix $\R_{\Q\G}$ and the translation vector $\t_{\Q\G}$) of a query image $\I_{\Q}$ coming from a calibrated camera.


Dense descriptors $\H_\Q$ of $\I_\Q$ are extracted using a convolutional neural
network $\mathcal{F}$ with parameters $\Theta$: $\H_\Q \defeq
\mathcal{F}\parr{\I_\Q;\Theta}$. Similarly, $\mathcal{F}$ is used to compute
a set of descriptors $\curl{\h_{n}}_{n=1...N}$ for each 3D point
$\left\{\u_{n}^{\G}\right\}_{n=1...N}$ in the database $\mathcal{D}$.

%

The warping function 
$\omega\!\parr{\u_{n}^{\G},\!\R_{\Q\G},\!\t_{\Q\G}}\!\!\defeq\!\!\K{\pi\parr{\R_{\Q\G}\u_{n}^{\G}\text{+}\,\t_{\Q\G}}}$
allows to warp a 3D point $\u_{n}^{\G}$ to obtain a 2D point $\p_{n}^{\Q}$ onto
the image plane of $\I_\Q$, \ie
$\p_{n}^{\Q}=\omega\parr{\u_{n}^{\G},\R_{\Q\G},\t_{\Q\G}}$, where $\K$ is the  camera
calibration matrix and $\pi\parr{\u}\defeq\croch{\u_x/\u_z,\u_y/\u_z}^\T$ is the
projection function.  


Let us now introduce the concept of \emph{correspondence map}.  In this paper, the
correspondence map $\C_{\Q,n}$ of $\u_{n}^{\G}$ in $\I_\Q$ is computed as follows:
$\C_{\Q,n} = g\parr{\h_{n} * \H_\Q}$ where $g$ is the softmax function and $*$ is the spatial convolution operator. 
The value $\C_{\Q,n}\parr{\p_{n}^{\Q}}$ describes how likely it is that pixel location $\p_{n}^{\Q}$ in $\I_\Q$ corresponds to $\u_{n}^{\G}$.  $\C_{\Q,n}$ also has an extra category $\p_{n}^{\Q}=\out$ that corresponds to the case where $\u_{n}^{\G}$ is not seen in $\I_\Q$.  By definition, $\C_{\Q,n}\parr{\p_{n}^{\Q}=\out}\defeq0$.  Thus, $\C_{\Q,n}$ has $|\mathring{\Omega}_\Q|=1+H_\Q \times W_\Q$ categories, where $H_\Q$ and $W_\Q$ are the number of rows and columns of $\H_\Q$, $\Omega_\Q$ is the set of all the pixel locations in $\H_\Q$ and $\mathring{\Omega}_\Q\defeq \left\{\Omega_\Q,\out \right\}$. 

The following notations will also be useful:
$\iverson{\cdot}$ is the Iverson bracket  ($\iverson{\text{True}} = 1$
  and $\iverson{\text{False}} = 0$), $\lfloor \cdot \rfloor$ is the floor function and $\norm{\cdot}$ is the L2
norm.


\begin{figure*}[!t]
  \centering
  \includegraphics[width=0.85\textwidth]{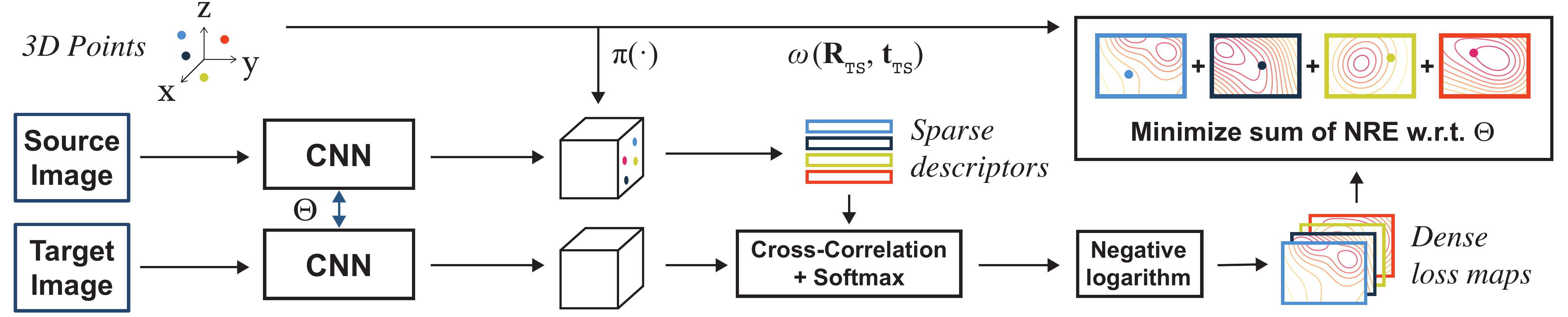}
  \caption{
    \textbf{Learning image descriptors tailored for camera pose estimation:} Given
    a pair of target/source images ($\I_\T,\,\I_\S$), 3D points
    $\curl{\u_{n}^{\S}} _{n=1...N}$ (seen in both $\I_\S$ and $\I_\T$) and
    ground truth camera poses ($\R_{\T\S}$ and $\t_{\T\S}$), we first extract
    dense representations for both images. For each 3D, we compute dense loss
    maps and minimize the S2D-CE loss with respect to the backbone parameters
    $\Theta$.
  }
  \label{fig:training}
\end{figure*}

\section{Neural reprojection error}\label{sec:NRE}
In this section, we first introduce the standard RE and then we present our novel NRE. 

\subsection{Reprojection error}\label{sec:RE}
The RE, that is used by most of the camera pose estimation methods, corresponds to the following equation:
\begin{equation}
\!\!\!\text{RE}\!\parr{\u_{n}^{\G},\p_n^{\Q},\R_{\Q\G},\t_{\Q\G}}\! \defeq \!\psi_\sigma\parr{\norm{\p_n^{\Q}\!-\!\omega\parr{\u_{n}^{\G},\R_{\Q\G},\t_{\Q\G}}}} \label{eq:standard_RE}
\end{equation}
where $\curl{\p_{n}^{\Q},\u_{n}^{\G}}$ is a 2D-3D correspondence and $\psi_\sigma\parr{\cdot}$ is a parametric robust loss, such as Geman-McClure or
Tukey’s biweight~\cite{zach2017iterated,barron2019general}, that allows to reduce the influence of large residuals.
Estimating the camera pose by minimizing a sum of RE terms enforces the 3D model and the query image to be summarized into a set of putative correspondences which results in a significant and irreversible loss of information.  This loss of information needs to be compensated as far as possible through the choice of a robust loss and the tuning of its hyperparameters, that usually depend on both the visual content and the outliers distribution. Moreover, outlier correspondences convey erroneous data to the pose estimator.
On the contrary, our novel loss, which we introduce in the next section, leverages richer information from the 3D model and the query image than RE and as a result eliminates the need for choosing a robust loss and its hyperparameters.

\subsection{Our novel loss}
Instead of computing the loss as a robust parametric function of the euclidean distance between the reprojected 3D point and its putative 2D correspondent in the query image, our novel loss function evaluates the discrepancy between two probability mass functions ({pmf}): the \emph{matching} {pmf} and the \emph{reprojection} {pmf}.  In the rest of this section, we first define these two {pmf} and then introduce our novel loss.

\begin{figure*}
  \begin{center}
    \centering
    \includegraphics[width=\textwidth]{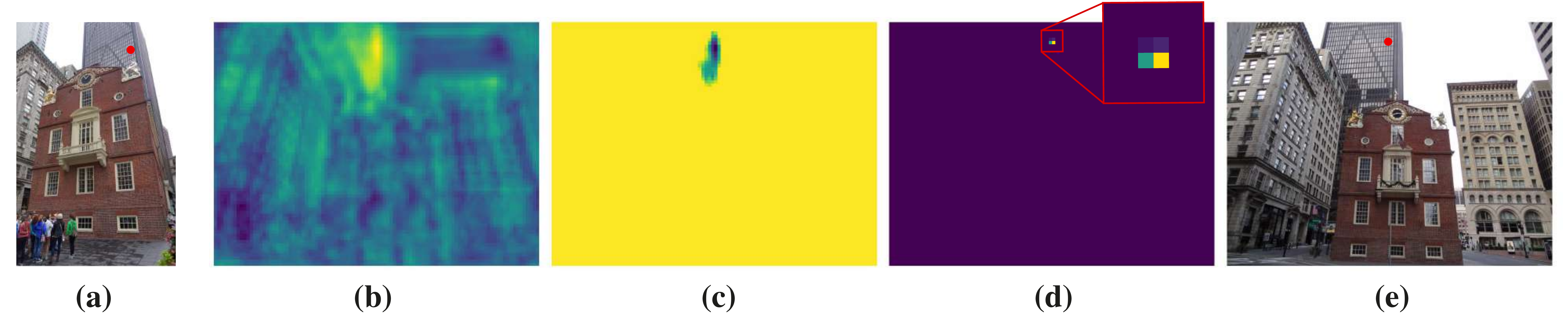}
    \caption{\textbf{Visualizations of maps involved in the derivation of NRE:} We show an example
      of \textbf{(a)} a source image and a reprojected 3D point $\u_{n}^{\G}$ using ground truth camera pose, \textbf{(b)} the non-robust dense loss map $\Ctilde_{\Q,n}$ with respect to the target image
      \textbf{(e)}, \textbf{(c)} the robust dense loss map $\L_{\Q,n}$, and
      \textbf{(d)} the target reprojection probability mass function used at
  training time.}
    \label{fig:maps}
    \vspace{-0.3cm}
  \end{center}
  \vspace{-0.3cm} 
\end{figure*}

\noindent\textbf{Matching probability mass function:}  This {pmf} describes how likely it is that the descriptor at the 2D image location $\p_{n}^{\Q}$ in $\H_\Q$ corresponds to the descriptor $\h_{n}$ of the 3D point $\u_{n}^{\G}$.
\begin{align}
  &q_\text{m} \parr{\p_n^\Q|s_n,\H_\Q,\h_{n}}\defeq
  s_n\,\C_{\Q,n}\parr{\p_n^\Q}+\frac{1-s_n}{|\mathring{\Omega}_\Q|} \> ,
  \label{eq:match_dist}
\end{align}
where  the binary  selector variable  $s_n\in\left\{ 0,\,1\right\}$  allows to
choose between two components: the  predicted correspondence map and the outlier
uniform {pmf}.  The latter component  introduces robustness
against  erroneous   correspondence   maps    that   may   occur   because   of
non-covisibility, occlusions, failure of the deep network, etc. We show in Fig.~\ref{fig:maps}(b) an example of the negative logarithm of a correspondence map.\\


\noindent\textbf{Reprojection probability mass function:}  This {pmf} describes how likely it is that a 2D location $\p_{n}^{\Q}\in \mathring{\Omega}_\Q$ corresponds to the reprojection of a 3D point $\u_{n}^{\G}$ using camera pose $\R_{\Q\G}$ and $\t_{\Q\G}$.
\begin{equation}
\begin{array}{l}
  q_\text{r}\left(\p_{n}^{\Q}|\u_{n}^{\G},\R_{\Q\G},\t_{\Q\G}\right)\defeq\\
  \quad\quad\mathtt{w}_{00,n}\iverson{\p_{n}^{\Q}=\left\lfloor
    \omega\parr{\u_{n}^{\G},\R_{\Q\G},\t_{\Q\G}}\right\rfloor}+\\
  \quad\quad\mathtt{w}_{10,n}\iverson{\p_{n}^{\Q}=\left\lfloor
    \omega\parr{\u_{n}^{\G},\R_{\Q\G},\t_{\Q\G}}\right\rfloor+\croch{1,0}^\T}+\\
  \quad\quad\mathtt{w}_{01,n}\iverson{\p_{n}^{\Q}=\left\lfloor
    \omega\parr{\u_{n}^{\G},\R_{\Q\G},\t_{\Q\G}}\right\rfloor+\croch{0,1}^\T}+\\
  \quad\quad\mathtt{w}_{11,n}\iverson{\p_{n}^{\Q}=\left\lfloor
    \omega\parr{\u_{n}^{\G},\R_{\Q\G},\t_{\Q\G}}\right\rfloor+\croch{1,1}^\T}\> ,
  \label{eq:reproj_dist}
\end{array}
\end{equation}
where the weights $\mathtt{w}_{i,j}$ are  bilinear interpolation coefficients,
\ie
\begin{equation}
  \begin{array}{ll}
    \mathtt{w}_{00,n}\defeq\parr{1-x_n}\parr{1-y_n}, &\mathtt{w}_{10,n}\defeq
    x_n\parr{1-y_n}, \\
    \mathtt{w}_{01,n}\defeq\parr{1-x_n} y_n, &\mathtt{w}_{11,n}\defeq x_n y_n,
  \end{array}\nonumber
\end{equation}
with
\begin{equation}
  \begin{array}{l}
  x_n\defeq\parr{\left\lfloor
    \omega\parr{\u_{n}^{\G},\R_{\Q\G},\t_{\Q\G}}\right\rfloor-
    \omega\parr{\u_{n}^{\G},\R_{\Q\G},\t_{\Q\G}}}_x,\\
y_n\defeq\parr{\left\lfloor
  \omega\parr{\u_{n}^{\G},\R_{\Q\G},\t_{\Q\G}}\right\rfloor-
  \omega\parr{\u_{n}^{\G},\R_{\Q\G},\t_{\Q\G}}}_y.
\end{array}\nonumber
\end{equation}
Equation~\ref{eq:reproj_dist}  sets a  non-zero weight  to the  four
image  locations   surrounding  the  reprojection  of   the  3D  point
$\u_{n}^{\G}$   under   camera   pose   parameters   $\R_{\Q\G}$   and
$\t_{\Q\G}$, and a zero weight to the rest of the image. In   a   slight   abuse   of    notation, if a reprojection $\omega\parr{\u_{n}^{\G},\R_{\Q\G},\t_{\Q\G}}$ falls outside of the image boundaries or if the 3D point has negative depth, \ie $\parr{\R_{\Q\G}\u_{n}^{\G}+\t_{\Q\G}}_z \leq 0$, we   consider   that
\begin{equation}
  \begin{array}{l}
    \left\lfloor \omega\parr{\u_{n}^{\G},\R_{\Q\G},\t_{\Q\G}}\right\rfloor+\croch{\cdot,\cdot}^\T \defeq\out\text{ and}\\
    \mathtt{w}_{00,n}\defeq 1, \; \mathtt{w}_{10,n}=\mathtt{w}_{01,n}=\mathtt{w}_{11,n}\defeq 0.
\end{array}\nonumber
\end{equation}
We show in Fig.~\ref{fig:maps}(d) an example of a reprojection {pmf}.

%

Assuming perfect descriptors and a perfect camera pose, the two {pmf} should be the same. This analysis is the fundamental idea of this paper: (a) given ground truth camera pose, we will make the \emph{matching} {pmf} fit the \emph{reprojection} {pmf} to learn descriptors tailored for pose estimation, (b) given descriptors, we will make the \emph{reprojection} {pmf} fit the \emph{matching} {pmf} to estimate the camera pose.

We propose to evaluate the discrepancy between  the \emph{matching} {pmf} (Eq.~\ref{eq:match_dist}) and the \emph{reprojection} {pmf} (Eq.~\ref{eq:reproj_dist}) using the following Cross-Entropy~(CE):
\begin{align}
  &\quad\text{CE}\parr{q_\text{r}
    \parr{\p_{n}^{\Q}|\u_{n}^{\G},\R_{\Q\G},\t_{\Q\G}}||q_\text{m}
    \parr{\p_{n}^{\Q}|s_n,\H_\Q,\h_{n}}}\nonumber \\
  &=\,\text{-}\!\!\!\sum_{\p_{n}^{\Q}\in\mathring{\Omega}_\Q}\!\!\! q_\text{r}\parr{\p_{n}^{\Q}|\u_{n}^{\G},\R_{\Q\G},\t_{\Q\G}}\ln\parr{q_\text{m}\parr{\p_{n}^{\Q}|s_n,\H_\Q,\h_{n}}}\nonumber\\
  &=\,s_n \Ctilde_{\Q,n}\parr{\omega\parr{\u_{n}^{\G},\R_{\Q\G},\t_{\Q\G}}}+ \parr{1-s_n}\ln{|\mathring{\Omega}_\Q|}\nonumber\\
  &\defeq \text{NRE} \parr{\u_{n}^{\G},\H_\Q,\h_{n},\R_{\Q\G},\t_{\Q\G},s_n} \label{eq:NRE}
\end{align}
where $\Ctilde_{\Q,n}\parr{\p}\defeq-\ln\parr{\C_{\Q,n}\parr{\p}}\, \forall\p\in\mathring{\Omega}_\Q$ is called a \emph{dense loss map}. The notation $\Ctilde_{\Q,n}\parr{\omega\parr{\u_{n}^{\G},\R_{\Q\G},\t_{\Q\G}}}$ corresponds to performing a bilinear interpolation at location $\omega\parr{\u_{n}^{\G},\R_{\Q\G},\t_{\Q\G}}$ in $\Ctilde_{\Q,n}$. 

From the point of view of the 3D point $\u_{n}^{\G}$, Eq.~\ref{eq:NRE} is a reprojection loss that depends on descriptors extracted by a convolutional neural
network $\mathcal{F}$ (see Sec.~\ref{sec:background}). Thus, we will refer to Eq.~\ref{eq:NRE} as the \emph{Neural Reprojection Error}.

From a practical point of view, given query dense descriptors $\H_\Q$ as well as 3D points and descriptors
$\curl{\u_{n}^{\G}, \h_{n}}_{n=1...N}$, it is possible to estimate the camera pose  by minimizing a sum of
NRE terms \wrt $\R_{\Q\G}$, $\t_{\Q\G}$ and $\curl{s_{n}}_{n=1...N}$ (see
Sec.~\ref{sec:test}). 
Here,
NRE relies on the dense loss maps directly which significantly reduces the amount of lost information compared to RE. Consequently, the need for choosing a robust loss and its hyperparameters is eliminated and all the information is kept available to estimate the camera pose.

Our novel NRE is differentiable not only \wrt to the camera pose but also \wrt the descriptors $\H_\Q$ and $\h_{n}$. Thus, providing ground-truth camera poses and minimizing NRE \wrt the descriptors yields a well-posed feature learning problem tailored for the pose estimation (see Sec.~\ref{sec:training}). NRE merges the feature learning problem and the camera pose estimation problem in a new way and allows to rethink the recent end-to-end feature metric pose refinement (see Sec.~\ref{sec:discussion-direct-RE}).

\section{Camera pose estimation}\label{sec:test}

Our novel NRE can be used to estimate the camera pose. Given a query image, from which query dense descriptors $\H_\Q$ are extracted, as well as 3D points and descriptors
$\curl{\u_{n}^{\G}, \h_{n}}_{n=1...N}$, we obtain a camera pose estimate 
%
by minimizing the following sum of NRE terms (Eq.~\ref{eq:NRE}) \wrt $\R_{\Q\G}$ and $\t_{\Q\G}$:
\begin{align}
  &\mathcal{L}\parr{\R_{\Q\G},\t_{\Q\G}}\nonumber\\
  &=\min_{s_1,s_2,...,s_N} \sum_{n=1}^N \text{NRE} \parr{\u_{n}^{\G},\H_{\Q},\h_{n},\R_{\Q\G},\t_{\Q\G},s_n}\nonumber\\
  &=\sum_{n=1}^N \min\parr{\ln{|\mathring{\Omega}_\Q|},\Ctilde_{\Q,n}\parr{\omega\parr{\u_{n}^{\G},\R_{\Q\G},\t_{\Q\G}}}}\label{eq:true_test_loss}\\
  &\approx \sum_{n=1}^N \L_{\Q,n}\parr{\omega\parr{\u_{n}^{\G},\R_{\Q\G},\t_{\Q\G}}}\label{eq:test_loss}
\end{align}
where the loss maps $\L_{\Q,n}$ are defined as follows:
\begin{align}
  \L_{\Q,n}\parr{\p}\defeq {\min\parr{\ln{|\mathring{\Omega}_\Q|},\Ctilde_{\Q,n}\parr{\p}}}\, \forall\p\in\mathring{\Omega}_\Q\>.\label{eq:L}
\end{align}
Instead of performing a bilinear interpolation in $\Ctilde_{\Q,n}$ followed by a truncation as in Eq.~\ref{eq:true_test_loss}, we apply a truncation to each element of $\Ctilde_{\Q,n}$ once (Eq.~\ref{eq:L}) and then perform a bilinear interpolation (Eq.~\ref{eq:test_loss}).  This approximation enables both a sparse storage of each loss map $\L_{\Q,n}$ and an efficient smoothing procedure (see Sec.~\ref{sec:refinement}).

Our loss function is robust against
outliers, since large values in $\Ctilde_{\Q,n}$ are truncated at $\ln{|\mathring{\Omega}_\Q|}$.   We show in Fig.~\ref{fig:maps}(c) an example of a robust \emph{dense loss map}  ($\L_{\Q,n}$).

Minimizing Eq.~\ref{eq:test_loss} is a non-convex optimization problem, thus we proceed in two steps: a sampling-based initialization step followed by gradient-based refinement step. 

\subsection{Initialization step}

To obtain an initial pose estimate, we employ an \emph{M-estimator SAmple
Consensus} approach (MSAC) \cite{torr2000mlesac}. The method is very similar to
a RANdom SAmple Consensus approach (RANSAC)~\cite{Fischler1981RandomSC}) but
does not require any user defined inlier/outlier threshold. Each iteration
consists of 1) randomly sampling 3 loss maps, 2) estimating a putative camera pose from these 3 loss maps and 3) evaluating Eq.~\ref{eq:test_loss} with that putative camera pose. 
Step 2 can be efficiently implemented using a standard P3P solver since:
\begin{align}
  &\argmin_{\R_{\Q\G}, \t_{\Q\G}} \sum_{n=1}^3 \L_{\Q,n}\parr{\omega\parr{\u_{n}^{\G},\R_{\Q\G},\t_{\Q\G}}}\nonumber\\
  =\,&\text{P3P}\parr{\curl{\u_{n}^{\G},\argmin_\p  \L_{\Q,n}\parr{\p}}_{n=1...3}}.\label{eq:p3p_loss}
\end{align}

\subsection{Refinement step}\label{sec:refinement}
Refining the initial camera pose remains a difficult optimization problem since each loss map in Eq.~\ref{eq:test_loss} may have plateaus and local minima (see Fig.~\ref{fig:teaser} middle and bottom rows) and the initial pose estimate may not be accurate enough for a gradient-based method to avoid a poor local minimum.

Thus, we employ a Graduated Non-Convexity approach
(GNC)~\cite{blake1987visual} that builds a sequence of successively
smoother (and therefore easier to optimize) approximations of the original loss
function. The optimization scheme consists of optimizing the sequence of loss functions, with the solution from the
previous objective used as starting point for the next one. However, Eq.~\ref{eq:test_loss} is not a standard robust
optimization problem~\cite{zach2018descending}. Therefore, we propose to apply a Gaussian-homotopy-like method
\cite{mobahi2015link} and consider the following smoothed version of the original loss function (a derivation of that
equation is given in the appendix):
\begin{align}
  &\!\!\breve{\mathcal{L}}_{\sigma} \parr{\R_{\Q\G},\t_{\Q\G}}\defeq \nonumber\\
  &\!\!\sum_{n=1}^N \!\sum_{\q\in\Gamma_{\Q,n}} \!\!\!\!\text{-}\!\parr{\ln|\mathring{\Omega}_\Q|\text{-}\L_{\Q,n}\parr{\q}} \!k_\sigma\!\parr{\norm{\q\text{-}\omega\parr{\u_{n}^{\G},\R_{\Q\G},\t_{\Q\G}}}} \!\! \label{eq:IRLS_cost}
\end{align}
where $k_\sigma\parr{\norm{\r}}\defeq\frac{1}{2\pi\sigma^2}e^{-{\frac
    {\norm{\r}^2}{2\sigma^2}}}$ is an
isotropic Gaussian kernel with standard variation $\sigma$ and $\Gamma_{\Q,n}$ is the set of pixel locations whose corresponding values
in $\Ctilde_{\Q,n}$ have not been truncated in Eq.~\ref{eq:L}. In Eq.~\ref{eq:IRLS_cost}, a large value of $\sigma$ leads to a highly
smoothed version of the original loss function while
a small value of $\sigma$ corresponds to a loss function that is very similar to
Eq.~\ref{eq:test_loss}. Therefore, in practice, we will start the
optimization with a value of $\sigma$ that is large enough, to avoid getting stuck in a poor local minimum and progressively decrease its value. Since Eq.~\ref{eq:IRLS_cost} is a standard robust optimization problem,  we employ an Iterated Reweighted Least Squares~(IRLS) approach to minimize each optimization problem within the GNC~\cite{blake1987visual} and use the stopping criterion proposed in \cite{zach2018descending}. 

\subsection{Coarse-to-fine strategy}\label{sec:coarse-to-fine}
From a practical point of view, the robustness and the accuracy of the camera
pose estimate directly depends on the loss maps, especially their resolution.
However, producing high resolution loss maps is an inefficient strategy: most of
the computational time would be spent computing cross-correlations in regions
distant from the true correspondent locations.  Instead, we propose a
coarse-to-fine strategy: we first estimate a coarse camera pose using
low-resolution loss maps and then refine it using local high-resolution ones.

For a given query image of size $H\times W\times 3$, we proceed as follows: 1)
Coarse dense descriptors of size $H/16 \times W/16 \times 1280$ are extracted
using a \emph{coarse} network ($\mathcal{F}_{\text{coarse}}$). 2) Low-resolution
loss maps of size $H/16 \times W/16$ are computed. 3) We run an MSAC~\cite{torr2000mlesac}+P3P to obtain an initial coarse pose estimate. 4) We apply a GNC~\cite{blake1987visual} procedure (still using low-resolution correspondence maps) to refine that initial coarse estimate. 5) Fine dense descriptors of size $H/2 \times W/2 \times 288$ are extracted using a \emph{fine} network ($\mathcal{F}_{\text{fine}}$). 6) Local high-resolution loss maps of size $64 \times 64$ are computed at the location of the reprojected 3D points using the coarse pose estimate. 7) We apply a GNC~\cite{blake1987visual} procedure starting from the coarse pose estimate to obtain our final pose estimate. 
Implementation details are provided in the appendix.

This coarse-to-fine strategy allows to obtain a camera pose estimate very
efficiently while significantly reducing the amount of required memory, since we
never compute or store any high-resolution loss map (see
Tab.~\ref{table:timings}).

\section{Learning image descriptors}\label{sec:training}
Our novel camera pose estimation method (see Sec.~\ref{sec:test}) essentially consists in minimizing a sum of NRE terms, \wrt the camera pose, assuming that the underlying descriptor extractor networks $\mathcal{F}_{\text{coarse}}$ and $\mathcal{F}_{\text{fine}}$ provide robust and discriminative descriptors. Therefore, we need to learn these networks. Let us recall that NRE (Eq.~\ref{eq:NRE}) is differentiable \wrt the descriptors $\H_\Q$ and $\h_{n}$. Thus we can learn to extract descriptors using NRE as training loss. 
We provide pairs of target/source images ($\I_\T,\,\I_\S$), 3D points $\curl{\u_{n}^{\S}} _{n=1...N}$ (seen in both $\I_\S$ and $\I_\T$) and ground truth camera poses ($\R_{\T\S}$ and $\t_{\T\S}$). For each pair of images, we perform gradient descent over the following loss function (see Fig.~\ref{fig:training}):
\begin{align}
  \mathcal{L}\parr{\Theta}&=\sum_{n=1}^N \text{NRE} \parr{\u_{n}^{\S},\H_{\T},\h_n,\R_{\T\S},\t_{\T\S},s_n=1},\label{eq:training_loss}
\end{align}
with $\H_\T\!=\!\mathcal{F}\parr{\I_\T;\Theta}$,
$\H_\S\!=\!\mathcal{F}\parr{\I_\S;\Theta}$ and
$\h_n\!=\!\H_\S\parr{\K{\pi\parr{\u_{n}^{\S}}}}$. The selector variable $s_n$ is
set to one in order to ease the gradient propagation. As explained in
Sec.~\ref{sec:coarse-to-fine}, in practice, we employ two networks: a
\emph{coarse} network $\mathcal{F}_{\text{coarse}}$ and a \emph{fine} network
$\mathcal{F}_{\text{fine}}$. Thus we need to train two networks with
different architectures, which are detailed in the appendix.

\begin{table*}[t]
  \ra{1.05}
  \begin{center}
    \resizebox{0.99\textwidth}{!}{
      \begin{tabular}{@{} llc *{15}{c@{\hskip0.2in}} @{}}\\
        \toprule
        	\makecell{Features}
        &\makecell{Pose estimator}
        &\makecell{Hyperparam.}
          &\multicolumn{3}{c}{Translation Error}
          &\multicolumn{3}{c}{Rotation Error}\\
        \cmidrule(lr){4-6}
        \cmidrule(lr){7-9}

        &&&\makecell{0.25m}
        &\makecell{1m}
        &\makecell{5m}
        &\makecell{2\degree}
        &\makecell{5\degree}
        &\makecell{10\degree}\\
        \midrule
        S2DNet~\cite{S2DNet}
        &RE \emph{LO-RANSAC}~\cite{Chum2003LocallyOR} &${\tau=4}$ & 0.54 (+23\%) & 0.45 (+32\%)  & 0.33 (+32\%) & 0.54 (+23\%) & 0.47 (+27\%) & 0.45 (+32\%)\\
        S2DNet~\cite{S2DNet}&RE \emph{GC-RANSAC}~\cite{barath2018graph} &${\tau=4}$& 0.54 (+23\%) & 0.43 (+26\%) & 0.31 (+24\%) & 0.53 (+20\%) & 0.47 (+27\%) & 0.43 (+26\%)\\
        S2DNet~\cite{S2DNet}&RE \emph{MAGSAC++~}\cite{MAGSACpp} &N/A& 0.51 (+16\%) & 0.43 (+26\%) & 0.31 (+24\%)  & 0.51 (+16\%) & 0.45 (+22\%) & 0.42 (+24\%)\\
        S2DNet~\cite{S2DNet}&RE \emph{Minimize Eq.~\ref{eq:RE_IRLS_cost}} &${\sigma=5}$ & 0.53 (+20\%) & 0.44 (+29\%) & 0.31 (+24\%) & 0.52 (+18\%) & 0.46 (+24\%) & 0.43 (+26\%)\\
        S2DNet~\cite{S2DNet} & FPR \emph{Minimize Eq.~\ref{eq:direct_method}} & Cf. Appendix & 0.49 {(+11\%)} & 0.42 {(+24\%)} & 0.30 {(+20\%)} & 0.48 {(+\phantom{0}9\%)} &
        0.44 {(+19\%)} & 0.42 {(+24\%)}\\
        S2DNet~\cite{S2DNet}&NRE &N/A& \textbf{0.44} {(+\phantom{0}0\%)} & \textbf{0.34} {(+\phantom{0}0\%)} & \textbf{0.25} {(+\phantom{0}0\%)} & \textbf{0.44} {(+\phantom{0}0\%)} & \textbf{0.37} {(+\phantom{0}0\%)} & \textbf{0.34} {(+\phantom{0}0\%)}\\
        \bottomrule
      \end{tabular}%
    }
  \end{center}
  \caption{
\textbf{NRE-based vs. RE-based pose estimators:} We evaluate the gain in performance of our novel NRE-based pose estimator against state-of-the-art RE-based pose estimators on the MegaDepth dataset~\cite{Megadepth}.  For a fair comparison, each method employs S2DNet~\cite{S2DNet} features, even our NRE-based pose estimator. For the methods that have an hyperparameter, we 
optimized it and report the best results. We report the error at several thresholds for
translation and rotation (lower is better). The scores between brackets show the relative deterioration \wrt to NRE. We
find that our NRE-based pose estimator significantly outperforms all the RE-based estimators.
We include the performance of the best feature-metric pose refinement estimator, which is covered in more details in the appendix.
\label{table:re_vs_nre}
}

\end{table*}

\section{Discussion}

\subsection{RE is a special case of NRE}\label{sec:discussion-RE}
In RE-based pose estimation, we are given 2D-3D correspondences $\curl{\u_{n}^{\G},\p_{n}^{\Q}}_{n=1...N}$. Let us consider a single 2D-3D correspondence. Assuming that $\p_{n}^{\Q}$ has integer pixel coordinates, we can build a one-hot-encoded correspondence map $\C_{\Q,n}$ such that $\C_{\Q,n}\parr{\p_{n}^{\Q}}=1$ and zeros everywhere else. In this case, Eq.~\ref{eq:L} is a dense loss map $\L_{\Q,n}$ that equals zero at the location $\p_{n}^{\Q}$ and $\ln|\mathring{\Omega}_\Q|$ everywhere else, and Eq.~\ref{eq:IRLS_cost} becomes:
\begin{equation}
  \!\!\!\mathcal{L}_{\sigma} \! \parr{\R_{\Q\G},\t_{\Q\G}} \! = \!\! \sum_{n=1}^N \!\!\text{-} \ln|\mathring{\Omega}_\Q| k_\sigma\!\parr{\norm{\p_n^{\Q}\text{-}\omega\parr{\u_{n}^{\G},\R_{\Q\G},\t_{\Q\G}}}}\!\!.\!\!\! \label{eq:RE_IRLS_cost}
\end{equation}
In Eq.~\ref{eq:RE_IRLS_cost}, each term within the sum corresponds to Eq.~\ref{eq:standard_RE} with a negative gaussian function as robust loss, whose shape is similar to the truncated quadratic kernel \cite{zach2017iterated}. Thus, RE is a special case of NRE. In the experiments, we will consider minimizing Eq.~\ref{eq:RE_IRLS_cost} to fairly compare RE \vs NRE.

\subsection{NRE \vs End-to-end feature metric pose refinement}\label{sec:discussion-direct-RE}
End-to-end Feature metric Pose Refinement (FPR) methods~\cite{ lv2019taking, tang2018ba, von2020gn} seek to minimize a loss of the following form at "test-time":
\begin{equation}
  \!\!\!\mathcal{L}_{\sigma}\parr{\R_{\Q\G},\t_{\Q\G}}\defeq\sum_{i=1}^N \!\psi_\sigma\!\parr{\norm{\h_{n} \text{-} \H_\Q \parr{\omega\parr{\u_{n}^{\G},\R_{\Q\G},\t_{\Q\G}}}}}\!.\!\!\!\label{eq:direct_method}
\end{equation}
In Eq.~\ref{eq:direct_method}, each term within the sum consists in reprojecting a 3D point into the query image plane but taking the distance in the space of descriptors. From this point of view, FPR is similar to NRE as it tries to leverage richer image information than simple 2D-3D correspondences. However FPR still requires choosing/learning a robust loss and tuning/learning its hyperparameters, so from this point of view it has the same limitations as RE.

But the major difference between FPR and NRE is that minimizing
Eq.~\ref{eq:direct_method} \wrt the descriptors does not yield a well-posed
feature learning problem. 
In order to learn descriptors tailored for pose estimation, FPR methods must
consider at least two losses. In~\cite{von2020gn}, a pixelwise contrastive loss is added (as well as a term involving the Hessian of the pose), while \cite{ lv2019taking} and \cite{tang2018ba} unroll several steps of an optimizer to obtain a computational graph and use a distance between the ground truth pose and the predicted pose to supervise the training.
On the contrary, minimizing NRE \wrt the descriptors yields a well-posed feature learning problem. Thus NRE is the first method to unify the feature learning problem and the camera pose estimation problem in a single loss and allows to rethink the end-to-end FPR strategy.

\begin{table*}[t]
\vspace{-0.3cm}
  \ra{1.05}
  \begin{center}
    \resizebox{0.95\textwidth}{!}{
      \begin{tabular}{@{} llc *{8}{c@{\hskip0.2in}} @{}}%
        \\
        \toprule
          \makecell{Category}
        &\makecell{Features}
        &\makecell{Pose estimator}&\multicolumn{3}{c}{Translation Error}
          &\multicolumn{3}{c}{Rotation Error}\\
        \cmidrule(lr){4-6}
        \cmidrule(lr){7-9}
	
        &&&\makecell{0.25m}
        &\makecell{1m}
        &\makecell{5m}
        &\makecell{2\degree}
        &\makecell{5\degree}
        &\makecell{10\degree}\\
        \midrule
	\multirow{2}{*}{Easy}
        
        &S2DNet& NRE & 0.17 (+\phantom{0}42\%) & 0.12 (+100\%) & 0.09 (+200\%) & 0.16 (+\phantom{0}45\%) & 0.13 (+\phantom{0}86\%) & 0.10 (+100\%)\\
        &NRE Features& NRE & \textbf{0.12} (+\phantom{00}0\%) & \textbf{0.06} (+\phantom{00}0\%) &
      \textbf{0.03} (+\phantom{00}0\%) & \textbf{0.11} (+\phantom{00}0\%) & \textbf{0.07} (+\phantom{00}0\%) & \textbf{0.05} (+\phantom{00}0\%)\\
        \midrule
        \multirow{2}{*}{Medium}
        &S2DNet& NRE & 0.29 (+\phantom{0}53\%) & 0.20 (+\phantom{0}67\%) & 0.15 (+150\%) & 0.27 (+\phantom{0}60\%) & 0.22 (+\phantom{0}69\%) & 0.19 (+\phantom{0}90\%)\\
        &NRE Features& NRE & \textbf{0.19} (+\phantom{00}0\%) & \textbf{0.12} (+\phantom{00}0\%) & \textbf{0.06} (+\phantom{00}0\%) & \textbf{0.17} (+\phantom{00}0\%) &
          \textbf{0.13} (+\phantom{00}0\%) & \textbf{0.10} (+\phantom{00}0\%) \\
        \midrule
       \multirow{2}{*}{Hard}
        &S2DNet& NRE & 0.44 (+\phantom{0}30\%) & 0.34 (+\phantom{0}42\%) & 0.25 (+108\%) & 0.44 (+\phantom{0}33\%) & 0.37 (+\phantom{0}37\%) & 0.34 (+\phantom{0}42\%)\\
        &NRE Features& NRE & \textbf{0.34} (+\phantom{00}0\%)  & \textbf{0.24} (+\phantom{00}0\%)  & \textbf{0.12} (+\phantom{00}0\%) & \textbf{0.33} (+\phantom{00}0\%)  &
        \textbf{0.27} (+\phantom{00}0\%)  & \textbf{0.24} (+\phantom{00}0\%)\\
        \bottomrule
      \end{tabular}%
    }
  \end{center}
  \caption{
    \textbf{NRE-based pose estimator using NRE features \vs NRE-based pose
    estimators using S2DNet features:} We evaluate the gain in performance of our NRE features against S2DNet~\cite{S2DNet} features using the same NRE-based pose estimator.
 We compare pose estimation on Megadepth~\cite{Megadepth} images evenly split in
 three difficulty categories.  We report the error at several thresholds for translation and rotation (lower is better). The scores between brackets show the relative deterioration \wrt to NRE features. We show that using our NRE features, the resulting estimated pose is markedly more accurate than using S2DNet features.
}
\label{table:nre_features_c2f}
\vspace{-0.4cm}

\end{table*}

\begin{table}
 \begin{center}
 \resizebox{0.99\columnwidth}{!}{
      \begin{tabular}{lcccc}%
        \toprule
        Features & S2DNet & S2DNet & NRE\\
        Pose estimator & RE & NRE & NRE\\
        \midrule
        Feature extraction & 28.2ms & 28.2ms &  N/A \\
        Feature extraction \emph{coarse} &  N/A &  N/A & 15.5ms\\
        Feature extraction \emph{fine} &  N/A &  N/A & 7.2ms\\
        Compute correspondence maps & 300ms & 300ms &  N/A & \\
        Compute \emph{coarse} correspondence maps &  N/A &  N/A & 8ms\\
        Compute \emph{local fine} correspondence maps & N/A &  N/A& 3ms\\
        Pose initialization (single iteration) & 0.9ms & 1.1ms & 1.1ms\\
         Pose refinement & 0.11s & 0.61s &  N/A\\
 	Pose refinement \emph{coarse}& N/A & N/A & 0.15s\\
 	Pose refinement \emph{fine}& N/A & N/A & 0.28s\\
	Total features memory& 2949MB & 2949MB & 591MB\\
	Total correspondence maps memory& 7680MB & 7680MB & 46MB\\
        \bottomrule
      \end{tabular}%
      }
  \end{center}
\caption{
    \textbf{Computational time and memory requirement study:} We report the
    average inference time on Megadepth~\cite{Megadepth} images with 1000 3D
    points. We show that our coarse-to-fine approach enables a much faster pose
    estimation and allows for larger scene scaling.
  }
  \vspace{-0.6cm}
  \label{table:timings}
\end{table}

\begin{figure}

 \begin{center}
    \centering
    \includegraphics[width=\columnwidth]{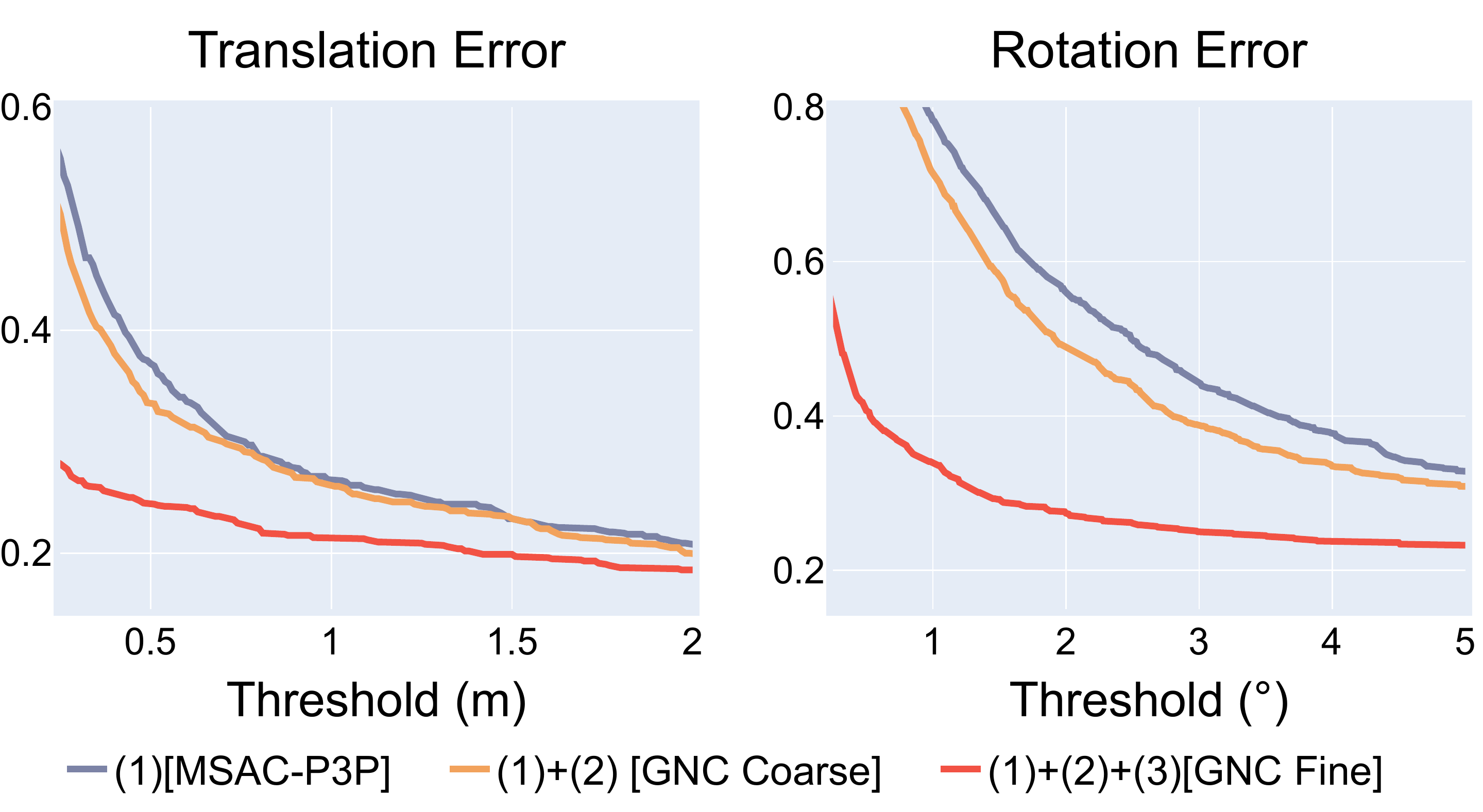}
    \caption{
      \textbf{Ablation study:} We report the cumulative error curves in
      pose estimation (lower is better), on the hardest
      category of our Megadepth study. We find that each step of our
      NRE-based coarse-to-fine estimator brings significant improvements.
    }
    \label{fig:ablation}
    \vspace{-0.5cm}
  \end{center}
\end{figure}

\section{Experiments}

In this section, we experimentally demonstrate that our novel NRE-based pose
estimator significantly outperforms state-of-the art RE-based pose estimators.
We also show that our coarse-to-fine strategy markedly reduces the amount of
required memory and the overall computational time of our NRE-based pose
estimator.

\subsection{Dataset and method}
We assembled an evaluation dataset of $3000$
Megadepth~\cite{Megadepth} image pairs, sampled from the validation set.  Using
the provided SfM model reconstructed using SIFT~\cite{SIFT}, we create image
pairs which contain at least 50 covisible 3D points. We evenly split them based
on their viewpoint distances to create three difficulty categories, which we
name \textit{Easy}, \textit{Medium} and \textit{Hard}.
At test-time for every pair of source and target images, we aim at predicting
the absolute camera pose of the target image, based on the 3D points visible in
the source image. We report the pose estimation error for several precision
thresholds.

\subsection{RE-based \vs NRE-based pose estimator}

In this first evaluation, we compare RE-based pose estimators against our novel
NRE-based pose estimator.  In order to have a fair comparison, we use
S2DNet~\cite{S2DNet} features for all methods evaluated in this study.

\noindent\textbf{Baselines:} We compare our NRE-based pose estimator against multiple
state-of-the-art RE-based pose estimators. This includes
LO-RANSAC~\cite{Chum2003LocallyOR}, GC-RANSAC~\cite{barath2018graph} and
MAGSAC++~\cite{MAGSACpp}, which all aim at finding inlier correspondences from
putative matches. We also add the minimization of Eq.~\ref{eq:RE_IRLS_cost} and Eq.~\ref{eq:direct_method}.

For all RE-based pose estimators, we follow S2DNet~\cite{S2DNet} and provide raw
putative 2D-to-3D matches based on the correspondence map argmax location. For
our NRE estimator, we use the same correspondence maps but preserve all the
information.
For all methods requiring hyperparameter tuning, we run several evaluations to
find the optimal one on our dataset. More details are provided in the appendix.

\noindent\textbf{Results:} We report pose estimation errors for the
aforementionned methods in Tab.~\ref{table:re_vs_nre}. We find our NRE-based pose
estimator consistently provides significant improvements over other RE-based
estimators.  In addition as shown in the appendix, we find 
hyperparameter tuning has a significant impact on performance for parametric RE
estimators. Our NRE-estimator however, requires no tuning.

\subsection{Coarse-to-fine experiment}

We provide an ablation study in Fig.~\ref{fig:ablation} of our coarse-to-fine
strategy. We find that each step of our NRE-based estimator brings significant
improvements.  We now compare the performance coupling the NRE estimator with
NRE features trained on the same training set as S2DNet~\cite{S2DNet}, using our
coarse-to-fine strategy. We report in Tab.~\ref{table:nre_features_c2f} the pose
estimation error on all categories from our Megadepth~\cite{Megadepth}
benchmark.
We find that using NRE features brings an additional leap in performance, by up
to $200\%$. Thanks to our coarse-to-fine formulation, this is all achieved at a
fraction of the cost of S2DNet~\cite{S2DNet}. As reported in
Tab.~\ref{table:timings}, NRE features have a memory footprint which is over
$16$ times lighter, while also performing a lot faster. This is a key component
for practical applications, or when scaling up to larger amount of keypoints or
images.
Additional qualitative and quantitative results are provided in the appendix.

\section{Conclusion} In this paper, we introduced the \emph{Neural Reprojection
Error} (NRE) as a substitute for the widely used Reprojection Error (RE). NRE
allows to perform absolute camera pose estimation by leveraging richer
information than RE and eliminates the need for choosing a robust loss and its
hyperparameters. We also proposed a coarse-to-fine optimization strategy that
allows to very efficiently minimize a sum of NRE terms \wrt the camera pose. We
experimentally demonstrated that replacing RE with NRE significantly improved
the accuracy and the robustness of the camera pose estimate while being
computationally and memory highly efficient. Our derivation of NRE merges the
feature learning problem and the absolute camera pose estimation problem in a
new way that allows to rethink the end-to-end feature-metric pose refinement
strategy. From a broader point of view, we believe this new way of merging deep
learning and 3D geometry may be useful in other computer vision applications.



\ifcase\HideSuppMat

\section*{Appendix}

In the following pages, we present additional quantitative results, qualitative
results and experimental details about the Neural Reprojection Error.

\appendix

\begin{figure*}[t]
  \begin{center}
    \centering
    \includegraphics[width=0.9\columnwidth]{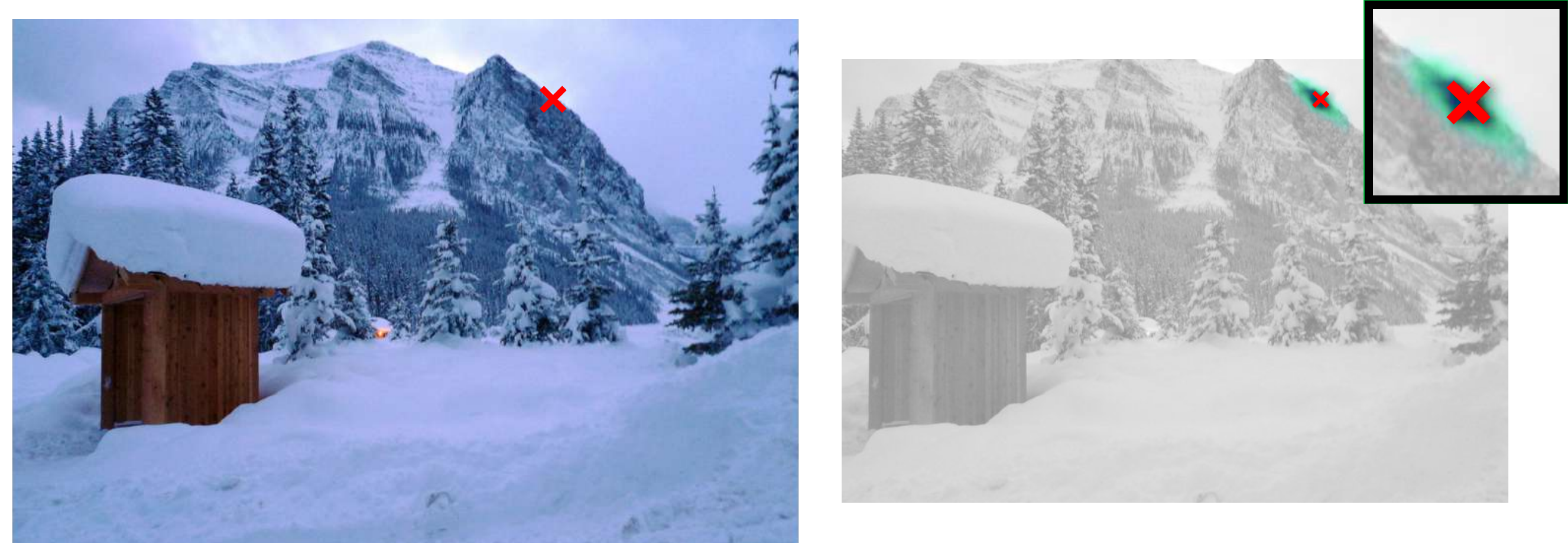}
    \includegraphics[width=0.9\columnwidth]{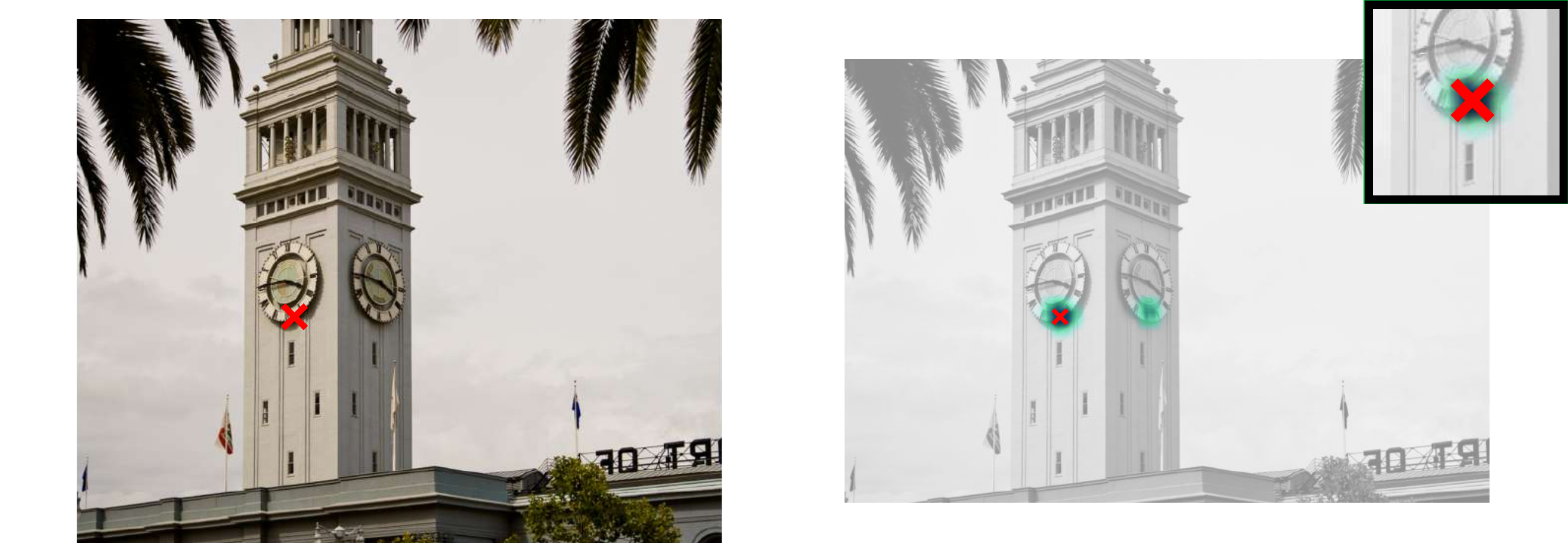}
    \includegraphics[width=0.9\columnwidth]{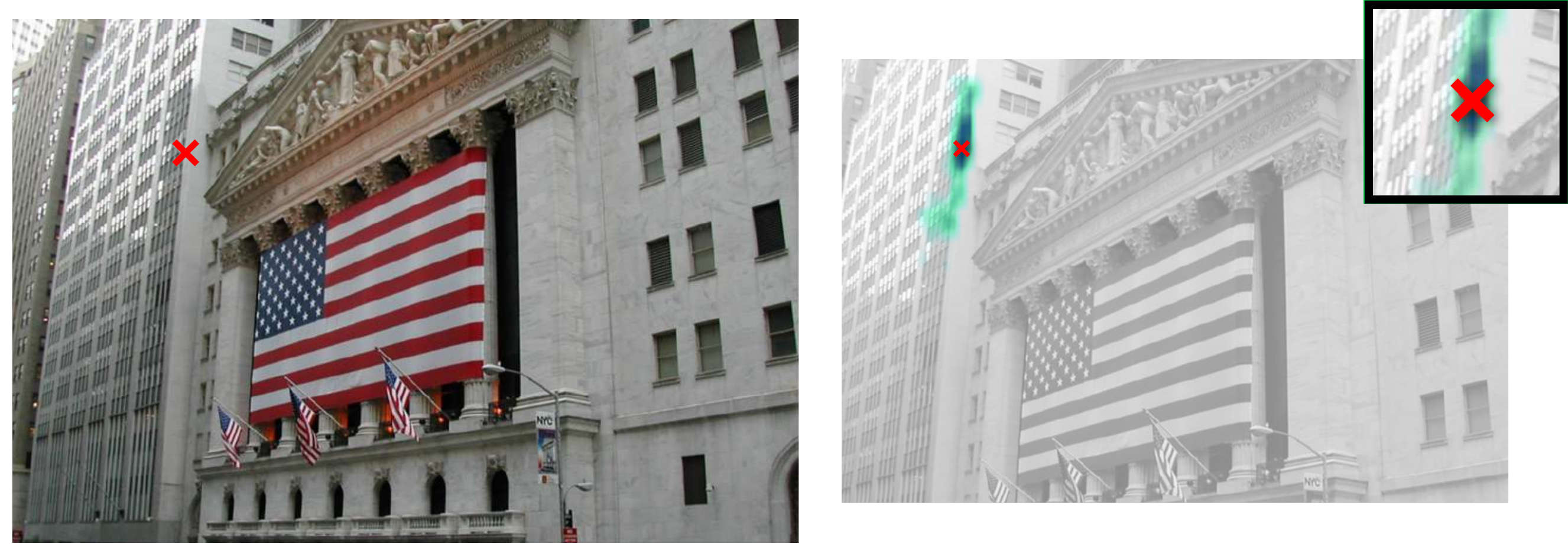}
    \includegraphics[width=0.9\columnwidth]{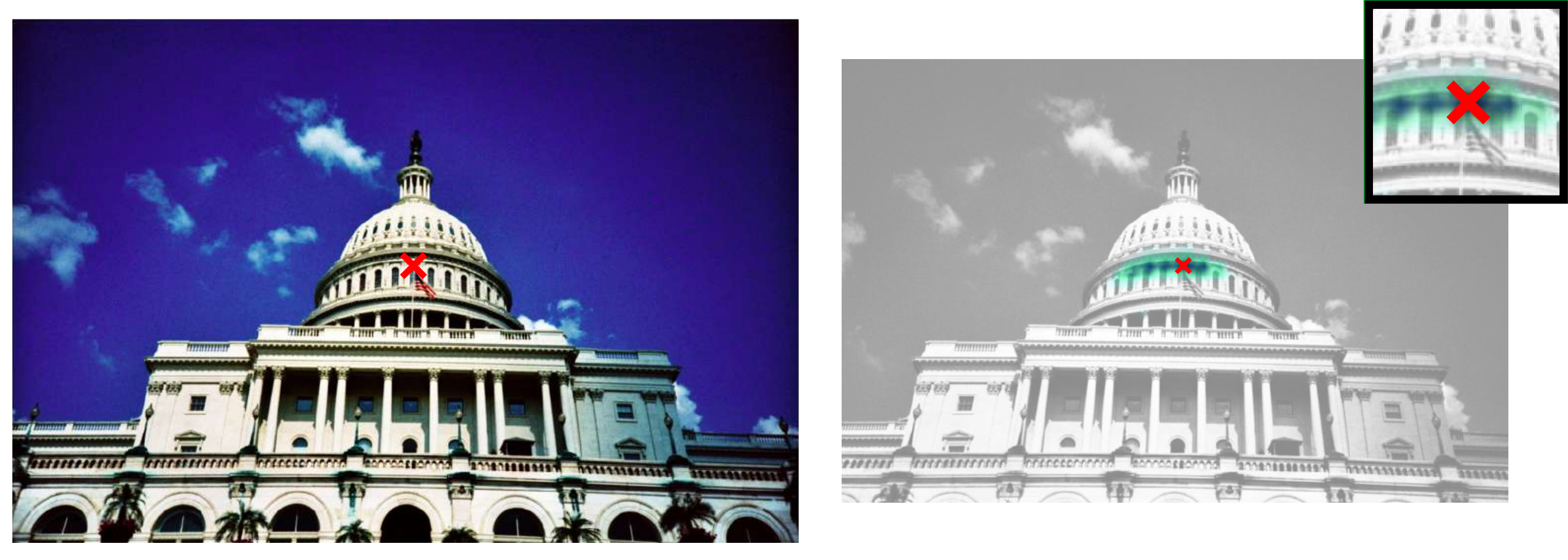}
    \vspace{0.5cm}
    \caption{
      \textbf{Qualitative results:} These qualitatives results correspond to additional examples for columns (a) and
      (b) in Fig.\ref{fig:teaser}.  It highlights that the dense loss maps keep much more information than RE. As a consequence our novel NRE-based pose estimator significantly outperforms RE-based pose estimators.
    }
    \label{fig:dense_loss_maps}
  \end{center}
\end{figure*}

\begin{table*}[h!]
  \ra{1.05}
  \begin{center}
    \resizebox{0.99\textwidth}{!}{
      \begin{tabular}{@{} llc *{16}{c@{\hskip0.2in}} @{}}\\
        \toprule
            \makecell{Features}
        &\makecell{Pose estimator}
        &\makecell{Fusion}
        &\makecell{$\psi$}
          &\multicolumn{3}{c}{Translation Error}
          &\multicolumn{3}{c}{Rotation Error}\\
        \cmidrule(lr){5-7}
        \cmidrule(lr){8-10}

         &&&&\makecell{0.25m}
        &\makecell{1m}
        &\makecell{5m}
        &\makecell{2\degree}
        &\makecell{5\degree}
        &\makecell{10\degree}\\
        \midrule
	S2DNet~\cite{S2DNet}&RE \emph{MAGSAC++~}\cite{MAGSACpp} &N/A& N/A&0.51 (+\phantom{0}16\%) & 0.43 (+\phantom{0}26\%) & 0.31 (+\phantom{0}24\%)  & 0.51 (+\phantom{0}16\%) & 0.45 (+\phantom{0}22\%) & 0.42 (+\phantom{0}24\%)\\
    S2DNet~\cite{S2DNet} & FPR \emph{Min. Eq.~\ref{eq:direct_method}} & C2F & Huber~\cite{Huber1964RobustEO} & 0.70 {(+\phantom{0}59\%)} & 0.65 {(+\phantom{0}91\%)} & 0.52
        {(+108\%)}& 0.69 {(+\phantom{0}57\%)}& 0.63 {(+\phantom{0}70\%)}& 0.58 {(+\phantom{0}71\%)}\\
        S2DNet~\cite{S2DNet} & FPR \emph{Min. Eq.~\ref{eq:direct_method}} & C2F &
        Barron~\cite{barron2019general} & 0.55 (+\phantom{0}25\%) & 0.44 (+\phantom{0}29\%) & 0.30 (+\phantom{0}20\%) & 0.55 (+\phantom{0}25\%) & 0.48 (+\phantom{0}30\%) & 0.43 (+\phantom{0}26\%)\\
        S2DNet~\cite{S2DNet} & FPR \emph{Min. Eq.~\ref{eq:direct_method}} & Concat. & Huber~\cite{Huber1964RobustEO} &
        0.49 {(+\phantom{0}11\%)} & 0.42 {(+\phantom{0}24\%)} & 0.30 {(+\phantom{0}20\%)} & 0.48 {(+\phantom{0}\phantom{0}9\%)} &
        0.44 {(+\phantom{0}19\%)} & 0.42 {(+\phantom{0}24\%)}\\
        S2DNet~\cite{S2DNet} & FPR \emph{Min. Eq.~\ref{eq:direct_method}} & Concat. &
         Barron~\cite{barron2019general} & 0.49 {(+\phantom{0}11\%)} & 0.42 {(+\phantom{0}24\%)} & 0.30 {(+\phantom{0}20\%)} & 0.48 {(+\phantom{0}\phantom{0}9\%)} &
        0.44 {(+\phantom{0}19\%)} & 0.42 {(+\phantom{0}24\%)}\\
        S2DNet~\cite{S2DNet}&NRE &N/A& N/A & \textbf{0.44} {(+\phantom{00}0\%)} & \textbf{0.34} {(+\phantom{00}0\%)} & \textbf{0.25} {(+\phantom{00}0\%)} & \textbf{0.44} {(+\phantom{00}0\%)} & \textbf{0.37} {(+\phantom{00}0\%)} & \textbf{0.34} {(+\phantom{00}0\%)}\\
        \bottomrule
      \end{tabular}%
    }
  \end{center}
  \caption{
    \textbf{NRE-based pose estimator \vs Feature-Metric Pose Refinement:} 
    We evaluate the gain in performance of our novel NRE-based pose estimator
    against the Feature-Metric Pose Estimation (FPR) variant on the MegaDepth dataset. Here FPR consists in
    minimizing Eq.~\ref{eq:direct_method} using as initialization the camera pose estimate from RE
    \emph{MAGSAC++~}\cite{MAGSACpp}. We find here that minimizing Eq.~\ref{eq:direct_method} allows to improve the camera pose estimate from \emph{MAGSAC++}, however our novel NRE again shows superior
    performance, while requiring no robust kernel selection. The scores between brackets show the relative deterioration w.r.t. to NRE.
  \vspace{0.2cm}
  \label{table:fpr_vs_nre}
  }
\end{table*}

\section{Additional Experiments}


\subsection{NRE-based pose estimator \vs Feature metric Pose Refinement}

We compare our novel NRE-based pose estimator against Feature-Metric Pose
Refinement (FPR) methods. As explained in Section~\ref{sec:discussion-direct-RE}, FPR methods seek to
minimize Eq.~\ref{eq:direct_method}. As such, FPR benefits from dense information contained in query
feature maps, but requires to choose a robust loss function and tune its hyperparameters. 

To complement our RE-based \vs NRE-based pose estimators study presented in Tab.~\ref{table:re_vs_nre}, we propose to reuse
S2DNet~\cite{S2DNet} features to perform FPR, initialized from our best RE pose
estimator (MAGSAC++~\cite{MAGSACpp}). To merge information from all three
feature extraction levels from S2DNet~\cite{S2DNet}, we try upsampling and
concatenating descriptors, as well as a coarse-to-fine alternative in which we
iteratively refine predictions from the previous (coarser) level.

We report pose estimation errors in Tab.~\ref{table:fpr_vs_nre} for FPR and NRE
estimators. We show results using the Huber~\cite{Huber1964RobustEO} robust
loss as well as the Barron~\cite{barron2019general} loss. We find that NRE
performs consistently better while eliminating the need for choosing a robust loss.

\subsection{Experiments on Aachen Night~\cite{6DOFBenchmark}}
So far, we evaluated the performances of our NRE-based pose estimator on MegaDepth~\cite{Megadepth}. Here, we run a similar study on the Aachen Night~\cite{6DOFBenchmark, Sattler2012ImageRF} dataset. This challenging outdoor
dataset consists of $4,328$ sparsely sampled daytime database images, and $98$
nighttime query images. To have a fair comparison between NRE-based and RE-based pose
estimators, we pair each query image with an oracle nearest-neighbor database
image 
and use all of its visible 3D points to predict the query pose.
Similar to the MegaDepth study, we report results for RE-based, FPR-based and
NRE-based pose estimators, using S2DNet features in Tab.~\ref{table:aachen}. For
FPR-based pose estimators we pick the best configuration
from~\ref{table:fpr_vs_nre}.

 As in the MegaDepth experiment, our NRE-based pose estimator consistently
 provides significant improvement over other pose estimators. We also compare
 the performance coupling the NRE-based pose estimator with NRE features trained
 on the same training set as S2DNet~\cite{S2DNet}. We report in
 Tab.~\ref{table:nre_features_c2f_appendix} the pose estimation errors. We again find
 that using NRE features brings an additional leap in performance.

\subsection{Experiments on InLoc~\cite{Taira2018InLocIV}}

To evaluate the generalization capabilities in an indoor scenario, we run the
same experiment on the InLoc~\cite{Taira2018InLocIV} dataset. This dataset
consists of $329$ query images, for $9,972$ database images. Unlike Aachen
Night, we have access to dense aligned depth maps for all database images. To
provide a fair comparison, we also pair each query image with an oracle
nearest-neighbor database image and use SuperPoint~\cite{SuperPoint} detections
(lifted to 3D using the depth maps) in the database images as inputs. Results
are reported in Tab.~\ref{table:aachen}.

We find that our NRE-based pose estimator provides consistent improvements at the
coarsest threshold, and overall competitive performance on the medium and fine
ones. The fact the relative improvement brought by our NRE-based pose estimator
is not as significant as for the other datasets can be attributed to the domain
shift with respect to the training images. Nonetheless, despite being trained on
outdoor images we find that our NRE features bring additional
improvements compared to S2DNet~\cite{S2DNet} features, as shown in
Tab.~\ref{table:nre_features_c2f_appendix}.

\section{Qualitative results}
In Fig.~\ref{fig:dense_loss_maps}, we show several examples of query images from the MegaDepth~\cite{Megadepth} validation set with a reprojected 3D point and the corresponding coarse dense loss map computed using our coarse NRE features. It highlights that the dense loss maps keep much more information than RE. As a consequence, as we show in our experiments, our novel NRE-based pose estimator significantly outperforms RE-based pose estimators.

\begin{table*}[t]
\vspace{0.3cm}
  \ra{1.05}
  \centering
\resizebox{\textwidth}{!}{ 
\begin{tabular}{@{} llc *{10}{c@{\hskip0.1in}} @{}}%
  \toprule
  Features & Pose Estimator & \multicolumn{3}{c}{Aachen Night} &
  \multicolumn{3}{c}{InLoc-DUC1} & \multicolumn{3}{c}{InLoc-DUC2}\\
  \cmidrule(lr){3-5}
  \cmidrule(lr){6-8}
  \cmidrule(lr){9-11}
  &&\makecell{0.25m, 2\degree}
  & \makecell{0.5m, 5\degree}
  & \makecell{5m, 10\degree}
  & \makecell{0.25m, 2\degree}
  & \makecell{0.5m, 5\degree}
  & \makecell{5m, 10\degree}
  & \makecell{0.25m, 2\degree}
  & \makecell{0.5m, 5\degree}
  & \makecell{5m, 10\degree}
  \\
  \midrule
  S2DNet & MAGSAC++~\cite{MAGSACpp} &
  0.46 (+\phantom{0}55\%) & 0.28 (+\phantom{0}80\%) & 0.10 (+229\%) &
  0.62 (+\phantom{00}3\%) & 0.41 (+\phantom{00}2\%) & 0.31 (+\phantom{0}11\%)&
  0.70 (+\phantom{0}11\%) & 0.44 (+\phantom{00}5\%) & 0.30 (+\phantom{00}2\%)
  \\
  S2DNet & RE \emph{Min. Eq.~10} &
  0.32 (+\phantom{00}7\%) & 0.20 (+\phantom{0}27\%) & 0.08 (+165\%) & 
  \textbf{0.58} (-\phantom{00}4\%) & 0.40 (+\phantom{00}1\%) & 0.31 (+\phantom{0}13\%) &
  0.66 (+\phantom{00}6\%) & 0.47 (+\phantom{0}13\%) & 0.39 (+\phantom{0}31\%)
  \\
  S2DNet & FPR \emph{Min. Eq.~11} &
  0.32 (+\phantom{00}7\%) & 0.20 (+\phantom{0}27\%) & 0.06 (+\phantom{0}97\%) &
  0.61 (+\phantom{00}1\%) & 0.41 (+\phantom{00}4\%) & 0.29 (+\phantom{00}4\%) &
  0.63 (+\phantom{00}1\%) & \textbf{0.41} (-\phantom{00}4\%) & 0.31 (+\phantom{00}5\%) &
  \\
  S2DNet & NRE &
  \textbf{0.30} (+\phantom{00}0\%) & \textbf{0.15} (+\phantom{00}0\%) & \textbf{0.03} (+\phantom{00}0\%) &
  0.60 (+\phantom{00}0\%) & \textbf{0.39} (+\phantom{00}0\%) & \textbf{0.28} (+\phantom{00}0\%) &
  \textbf{0.62} (+\phantom{00}0\%) & 0.42 (+\phantom{00}0\%) & \textbf{0.29} (+\phantom{00}0\%) &
  \\
  \bottomrule
  \end{tabular}%
}
\vspace{0.4cm}
\caption{\textbf{NRE-based \vs RE-based \vs FPR-based pose estimators on  Aachen
  Night~\cite{6DOFBenchmark} and InLoc~\cite{Taira2018InLocIV}:} We evaluate the gain in performance of our novel NRE-based pose estimator against
state-of-the-art RE-based and FPR-based pose estimators.  For a fair comparison, \emph{each method uses the same oracle nearest-neighbor database image for each query image}. Moreover, each method employs S2DNet~\cite{S2DNet} 
features, even our NRE-based pose estimator. For the methods that have an hyperparameter, we optimized it and report the best results. We
report the error at several thresholds for translation and rotation (lower is better). The scores between brackets show
the relative deterioration \wrt to NRE. On Aachen, there is no strong domain shift \wrt MegaDepth images that are used to train S2DNet, as a result the dense loss maps are accurate and our NRE-based pose estimator significantly outperforms its competitors. On InLoc, there is a strong domain shift (InLoc is an indoor dataset), as a result the dense loss maps are not very informative and our NRE-based pose estimator does not significantly outperform its competitors.
\label{table:aachen}}
\end{table*}
\begin{table*}[!h]
\vspace{0.4cm}
  \ra{1.05}
  \begin{center}
    \resizebox{\textwidth}{!}{
  \begin{tabular}{@{} llc *{10}{c@{\hskip0.2in}} @{}}%
    \toprule
    Features & \multirow{2}{*}{\makecell{Pose\\Estim.}} & \multicolumn{3}{c}{Aachen Night} &
    \multicolumn{3}{c}{InLoc-DUC1} & \multicolumn{3}{c}{InLoc-DUC2}\\
    \cmidrule(lr){3-5}
    \cmidrule(lr){6-8}
    \cmidrule(lr){9-11}
    &&\makecell{0.25m, 2\degree}
    & \makecell{0.5m, 5\degree}
    & \makecell{5m, 10\degree}
    & \makecell{0.25m, 2\degree}
    & \makecell{0.5m, 5\degree}
    & \makecell{5m, 10\degree}
    & \makecell{0.25m, 2\degree}
    & \makecell{0.5m, 5\degree}
    & \makecell{5m, 10\degree}
    \\
 	 \midrule
    S2DNet & NRE &
    0.30 (+\phantom{0}12\%) & 0.15 (+\phantom{0}37\%) & 0.03 (+\phantom{0}55\%) &
    0.60 (+\phantom{00}1\%) & 0.40 (+\phantom{00}3\%) & 0.28 (+\phantom{0}10\%) &
    0.63 (+\phantom{00}1\%) & 0.42 (+\phantom{0}10\%) & 0.30 (+\phantom{00}3\%) \\
    NRE Features & NRE &
    \textbf{0.26} (+\phantom{00}0\%) & \textbf{0.11} (+\phantom{00}0\%) & \textbf{0.02} (+\phantom{00}0\%)&
  \textbf{0.59} (+\phantom{00}0\%) & \textbf{0.39} (+\phantom{00}0\%) & \textbf{0.25} (+\phantom{00}0\%) &
    \textbf{0.62} (+\phantom{00}0\%) & \textbf{0.38} (+\phantom{00}0\%) &
    \textbf{0.29} (+\phantom{00}0\%) \\
  \bottomrule
  \end{tabular}%
}
  \end{center}
  \caption{
    \textbf{NRE features \vs S2DNet features for NRE-based pose estimators on Aachen Night~\cite{6DOFBenchmark} and
    InLoc~\cite{Taira2018InLocIV}:} We evaluate the gain in performance of our NRE
    features against S2DNet~\cite{S2DNet} features using the same NRE-based pose
    estimator. We compare pose estimation on Aachen Night~\cite{6DOFBenchmark} and
    InLoc~\cite{Taira2018InLocIV} images.  For a fair comparison, \emph{each
    method uses the same oracle nearest-neighbor database image for each query
    image}. We report the error at several precision thresholds for
    translation and rotation (lower is better). The scores between brackets show the
    relative deterioration \wrt to NRE features. On Aachen, there is no strong domain shift \wrt MegaDepth images that are used to train both S2DNet and our NRE feature, as a result the dense loss maps are accurate and we obtain improvements similar to the ones we obtained in our MegaDepth experiment. On InLoc, there is a strong domain shift (InLoc is an indoor dataset), as a result neither S2DNet dense loss maps nor the dense loss maps obtained using our NRE features are very informative. As a result,  the pose estimated ugin NRE features is not markedly more accurate than the pose obtained
    using S2DNet features.
  }
\label{table:nre_features_c2f_appendix}
\end{table*}

\begin{figure*}
  \begin{center}
    \centering
    \includegraphics[width=0.9\textwidth]{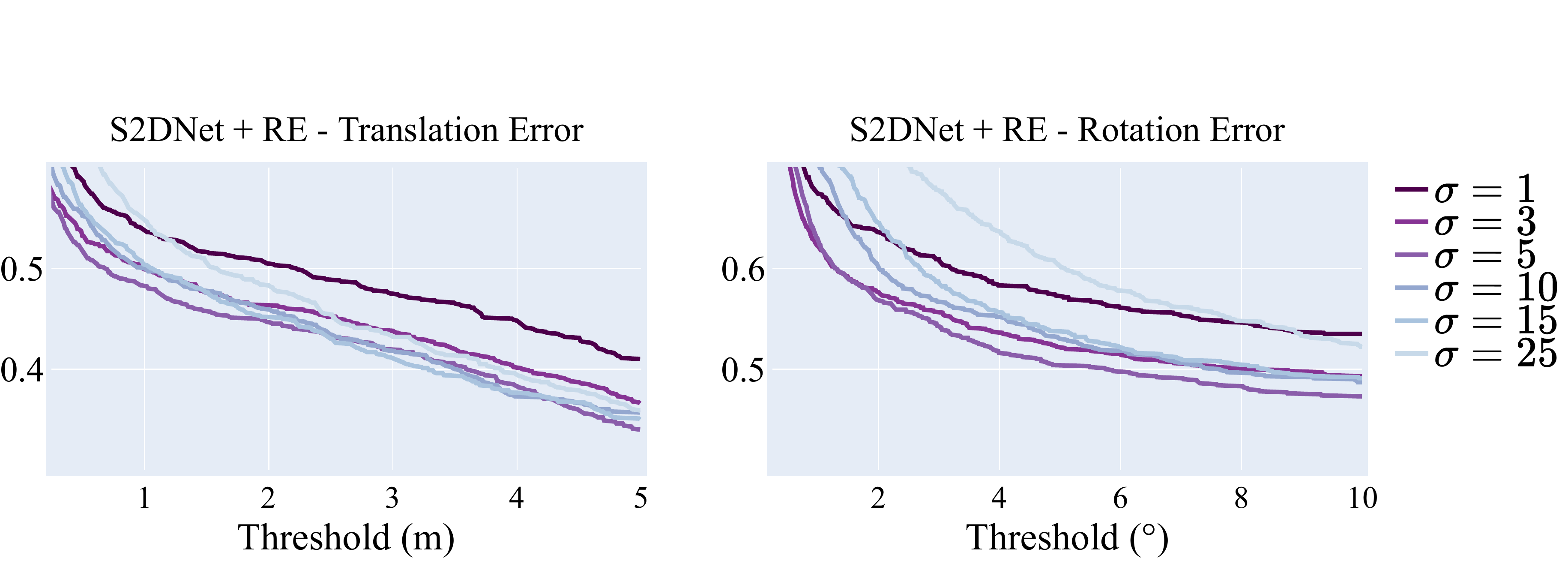}
    \caption{
      \textbf{Tuning the hyperparameter of an RE-based pose estimator:} We report the cumulative error curves in pose estimation (lower is better), on the hardest category of our Megadepth study, for the RE-based pose estimator that consists in minimize Eq.~$10$. We find that a careful hyperparameter tuning is very important. On the contrary, our novel formalism leads to a loss that does not possess any hyperparameter.
    }
    \label{fig:sigmas}
    \vspace{-0.1cm}
  \end{center}
\end{figure*}

\section{Derivation of Equation~\ref{eq:IRLS_cost}}
In this section, we show how Eq.~$8$ (in the submited wersion of the paper) is obtained.

The robust dense loss map $\L_{\Q,n,\sigma}$ can be smoothed using an isotropic Gaussian kernel as follows:

\begin{align}
&\breve\L_{\Q,n,\sigma}\parr{\p}\defeq {\sum_{\r} k_\sigma\parr{\norm{\r}} \L_{\Q,n}\parr{\p+\r}}\nonumber \\
= & \, \L_{\Q,n}\parr{\out}{\sum_{\r} k_\sigma\parr{\norm{\r}}}\nonumber \\
&+\sum_{\r} k_\sigma\parr{\norm{\r}} \parr{\L_{\Q,n}\parr{\p+\r}-\L_{\Q,n}\parr{\out}} \\
= & \, \sum_{\r} k_\sigma\parr{\norm{\r}} \parr{\L_{\Q,n}\parr{\p+\r}-\L_{\Q,n}\parr{\out}} + \text{cst}_\p\\
=&\sum_{\q\in\Omega_\Q} k_\sigma\parr{\norm{\q-\p}}
\parr{\L_{\Q,n}\parr{\q}-\L_{\Q,n}\parr{\out}} + \text{cst}_\p
\end{align}
\begin{align}
=&\sum_{\q\in\Omega_\Q} k_\sigma\parr{\norm{\q-\p}} \parr{\L_{\Q,n}\parr{\q}-\ln|\mathring{\Omega}_\Q|} + \text{cst}_\p \\
=&\sum_{\q\in\Gamma_{\Q,n}} k_\sigma\parr{\norm{\q-\p}} \parr{\L_{\Q,n}\parr{\q}-\ln|\mathring{\Omega}_\Q|} + \text{cst}_\p \label{eq:smoothed-cost-map}
\end{align}
where $k_\sigma\parr{\norm{\r}}\defeq\frac{1}{2\pi\sigma^2}e^{-{\frac{\norm{\r}^2}{2\sigma^2}}}$ is an
isotropic Gaussian kernel with standard variation $\sigma$ and $\Gamma_{\Q,n}$ is the set of pixel locations whose corresponding values
in $\L_{\Q,n}$ are lower than $\ln|\mathring{\Omega}_\Q|$.
Equation~\ref{eq:smoothed-cost-map} leads to the smoothed cost function:
\begin{align}
  &\!\!\breve{\mathcal{L}}_{\sigma} \parr{\R_{\Q\G},\t_{\Q\G}}\defeq \nonumber\\
  &\!\!\sum_{n=1}^N \!\sum_{\q\in\Gamma_{\Q,n}} \!\!\!\!\text{-}\!\parr{\ln|\mathring{\Omega}_\Q|\text{-}\L_{\Q,n}\parr{\q}} \!k_\sigma\!\parr{\norm{\q\text{-}\omega\parr{\u_{n}^{\G},\R_{\Q\G},\t_{\Q\G}}}}, \!\! 
\end{align}
which is a robust non-linear least squares problem and therefore can be minimized using the IRLS algorithm.

\section{Technical details}
\subsection{Coarse-to-Fine Strategy (Sec.~\ref{sec:coarse-to-fine})}
Step 6 of our coarse-to-fine strategy consists in computing local high-resolution loss maps of size $64 \times 64$ at the location of the reprojected 3D points using the coarse pose estimate. 
The idea of that step is to transform the low-resolution loss maps
into high-resolution loss maps to obtain a much more accurate pose
estimate. The question is: How can we combine a low-resolution robust loss map with a local high-resolution discriminative loss map ? We proceed as follows:
\begin{enumerate}
\item A coarse correspondence map  $\C_{\text{coarse}}$ is of size $H/16 \times W/16$. Let us recall that by definition $\sum_{\p\in\mathring{\Omega}_\text{coarse}} \C_{\text{coarse}}\parr{\p}=1$.
\item Compute the local high resolution correspondence map $\C_{\text{fine}}$ of size $64\times64$ at the location of the reprojected 3D points (using the coarse pose estimate) $\q$:
	\begin{enumerate} 
\item  Extract a $64\times64$ region in the dense fine descriptors around $\q$. 
\item Compute the dot product with the fine descriptor of the 3D point and apply a softmax to obtain $\C_{\text{fine}}$.
\end{enumerate} 
Thus by definition  $\sum_{\p\in\mathcal{N}_{64\times64}\parr{\q}} \C_{\text{fine}}\parr{\p}=1$.

\item $\C_{\text{fine}}$ corresponds to a region of size 8x8 in $\C_{\text{coarse}}$. Compute the sum of these 64 pixels in $\C_{\text{coarse}}$. We call this scalar $norm_{coarse}$.
\item Multiply $\C_{\text{fine}}$ by $\frac{norm_{coarse}}{64}$ to obtain $\C_{\text{fine norm}}$. $\C_{\text{fine norm}}$ is a local high-resolution version of $\C_{\text{coarse}}$.
\item The final local high resolution loss map is obtained classically: \\$\L_{\text{fine}}=\min\parr{\ln|\mathring{\Omega}_\text{fine}|,-\ln\parr{\C_{\text{fine norm}}}}$. By definition, outside of the $64\times64$ region, the value of the loss is $\ln|\mathring{\Omega}_\text{fine}|$.
\end{enumerate}

\subsection{Network Architectures (Sec.~\ref{sec:training})}

\noindent\textbf{Coarse network architecture.} The purpose of the coarse network
$\mathcal{F}_{\text{coarse}}$ is to provide robust descriptors that are used to
obtain a coarse pose estimate. To deal with ambiguous cases, it should leverage
image context. This motivates a deep architecture with a wide receptive field
and a large descriptor size. On the other hand, the network should output dense
descriptors of sufficient resolution to reliably estimate a coarse camera pose.
We experimentally found that an effective stride of 16 is sufficient. To satisfy
these specifications, we opted for an Inception-v3~\cite{Inceptionv3} backbone
and modified it accordingly. We changed some kernel sizes and truncated the
network at the layer Mixed-6e. In the end our final architecture has a receptive
field of $927$ pixels and produces dense descriptors of size  $H/16 \times W/16
\times 1280$.

\noindent\textbf{Fine network architecture.} The purpose of the fine network
$\mathcal{F}_{\text{fine}}$ is to provide discriminative high-resolution
descriptors that are used to refine the coarse pose estimate. However, producing
high-resolution descriptors takes a lot of memory. This motivates a deep
architecture with a small receptive field and a small descriptor size.  We
experimentally found that an effective stride of 2 is a good balance between
accuracy and memory consumption.  To satisfy these specifications, we opted
again for a modified Inception-v3~\cite{Inceptionv3} backbone. We only keep the
stride of 2 at the first layer and remove any Max-Pooling layer, and we truncate
the model at the Mixed-5d layer.  Our final architecture has a receptive field
of $43$ pixels and produces dense descriptors of size $H/2 \times W/2 \times
288$.\\

\noindent\textbf{Implementation details.} The coarse network
$\mathcal{F}_{\text{coarse}}$ and the fine network $\mathcal{F}_{\text{fine}}$
are trained independently. Both networks use the same training data which comes
from the MegaDepth dataset~\cite{Megadepth}. As D2-Net~\cite{D2Net}, we remove
scenes which overlap with the
PhotoTourism~\cite{Imc_phototourism,Thomee2016YFCC100MTN} test set. We train our
networks on image pairs ($\I_\S$ and $\I_\T$) with an arbitrary overlap. 

To train $\mathcal{F}_{\text{fine}}$, we extract random crops of size
$800\times800$ and randomly sample a maximum of 64 3D points visible in both
$\I_\S$ and $\I_\T$.
Using such large crops may seem an overkill since $\mathcal{F}_{\text{fine}}$
has a small receptive field. Let us highlight that using $C\times C$ crops
allows to produce correspondence maps of size $C/2\times C/2$ which essentially
consists in comparing each source patch against $C^2$ target patches.  Thus,
even if $\mathcal{F}_{\text{fine}}$ has a small receptive field,  the larger the
crops during training the better the descriptors, and $800\times800$ is the
maximum size that could fit in memory.

To train $\mathcal{F}_{\text{coarse}}$, we use entire images as inputs since the network has a very large receptive field and randomly sample a maximum of 64 3D points visible in both $\I_\S$ and $\I_\T$.
Each network is trained using early stopping on the MegaDepth validation set. We
use Adam~\cite{Kingma2014AdamAM} with an initial learning rate of $10^{−3}$ and apply a
multiplicative decaying factor of $e^{−0.1}$ at every epoch. 

\subsection{Timing}

We run all our training and experiments on a machine equipped with an Intel(R)
Xeon(R) E5-2630 CPU at 2.20GHz, and an NVIDIA GeForce GTX 1080Ti GPU.  The
timing results reported in Tab.~{table:timings} where obtained using
a Python implementation of the previously described algorithms. Source code will
be made available.

\subsection{Implementation details about the RE-based \vs NRE-based pose estimators study} 
\begin{itemize}

\item In our RE-based \vs NRE-based pose estimators study, we used
  LO-RANSAC~\cite{Chum2003LocallyOR}, GC-RANSAC~\cite{barath2018graph} and
  MAGSAC++~\cite{MAGSACpp} implementations provided in OpenCV
  4.5.0~\footnote{\scriptsize{\url{https://docs.opencv.org/master/d9/d0c/group__calib3d.html}}}.

\item We show in Fig.~\ref{fig:sigmas} the cumulative errors curves for several
  $\sigma$ values when minimizing Eq.~\ref{eq:standard_RE} on the hardest category of our
  Megadepth~\cite{Megadepth} study. These results stress how important a careful
  hyperparameter tuning is in standard RE pose estimators.

\item Throughout our paper we run the coarse GNC with decreasing $\sigma$ values
  ranging from $2.0$ to $0.6$. For the fine GNC, we use values between $8.0$ and
  $0.6$.

\end{itemize}

\fi

\section*{Acknowledgement}

This project has received funding from the Bosch Research Foundation (\textit{Bosch
Forschungsstiftung}).
\newpage

{
  \small
  \bibliographystyle{ieee}
  \bibliography{string,new} 
}

\end{document}